\documentclass[10pt,twocolumn,letterpaper]{article}

\usepackage[pagenumbers]{cvpr} 

\usepackage{graphicx}
\usepackage{multicol}
\usepackage{multirow}
\usepackage{float}
\usepackage{tcolorbox}
\usepackage{enumitem}
\usepackage{listings}
\usepackage{csquotes}

\definecolor{cvprblue}{rgb}{0.21,0.49,0.74}
\usepackage[pagebackref,breaklinks,colorlinks,allcolors=cvprblue]{hyperref}

\def\paperID{***} 
\def\confName{CVPR}
\def\confYear{2026}

\title{Attention in Space: Functional Roles of VLM Heads for Spatial Reasoning}

\author{\textbf{Xueqi Ma\textsuperscript{1}}\thanks{Both authors contributed equally to this research. } \quad
\textbf{Shuo Yang\textsuperscript{1}}\footnotemark[1] \quad
\textbf{Yanbei Jiang\textsuperscript{1}} \quad
\textbf{Shu Liu\textsuperscript{1}} \quad
\textbf{Zhenzhen Liu\textsuperscript{2}} \\
\textbf{Jiayang Ao\textsuperscript{1}} \quad
\textbf{Xingjun Ma\textsuperscript{3}} \quad 
\textbf{Sarah Monazam Erfani \textsuperscript{1}} \quad
\textbf{James Bailey\textsuperscript{1,4}}  \\
\textsuperscript{1}The University of Melbourne \quad
\textsuperscript{2}Australian National University \\
\textsuperscript{3}Fudan University \quad
\textsuperscript{4}Monash University \\
\texttt{\{xueqim1, shuo.yang.3, sarah.erfani, baileyj\}@unimelb.edu.au} \\
\texttt{\{yanbeij, shu6, jiayang.ao\}@student.unimelb.edu.au} \\
\texttt{zhenzhen1108@gmail.com} \quad
\texttt{xingjunma@fudan.edu.cn} \\
\texttt{james.a.bailey@monash.edu}
}

\begin{document}
\maketitle
\begin{abstract}
Despite remarkable advances in large Vision-Language Models (VLMs), spatial reasoning remains a persistent challenge.
In this work, we investigate how attention heads within VLMs contribute to spatial reasoning by analyzing their functional roles through a mechanistic interpretability lens. We introduce CogVSR, a dataset that decomposes complex spatial reasoning questions into step-by-step subquestions designed to simulate human-like reasoning via a chain-of-thought paradigm, with each subquestion linked to specific cognitive functions such as spatial perception or relational reasoning. Building on CogVSR, we develop a probing framework to identify and characterize attention heads specialized for these functions. Our analysis across diverse VLM families reveals that these functional heads are universally sparse, vary in number and distribution across functions. Notably, spatially specialized heads are fewer than those for other cognitive functions, highlighting their scarcity.
We propose methods to activate latent spatial heads, improving spatial understanding. Intervention experiments further demonstrate their critical role in spatial reasoning: removing functional heads leads to performance degradation, while emphasizing them enhances accuracy. This study provides new interpretability driven insights into how VLMs attend to space and paves the way for enhancing complex spatial reasoning in multimodal models.
\end{abstract}

\section{Introduction}

Spatial reasoning \cite{yang2025thinking,zhang2025look,yin2025spatial}, the ability to understand and reason about spatial relationships between objects, is a fundamental aspect of human cognition. It underlies our capacity to interpret scenes, describe object relations, and perform actions in complex environments. Despite remarkable progress in multimodal understanding, Vision-Language Models (VLMs) \citep{zhu2023minigpt,liu2023visual,lu2024deepseek} still struggle with spatial reasoning, even in simple orientation problems such as \enquote{Is the dog facing the horse?} 

\begin{figure*}[t]
  \centering
  \includegraphics[width=0.85\linewidth]{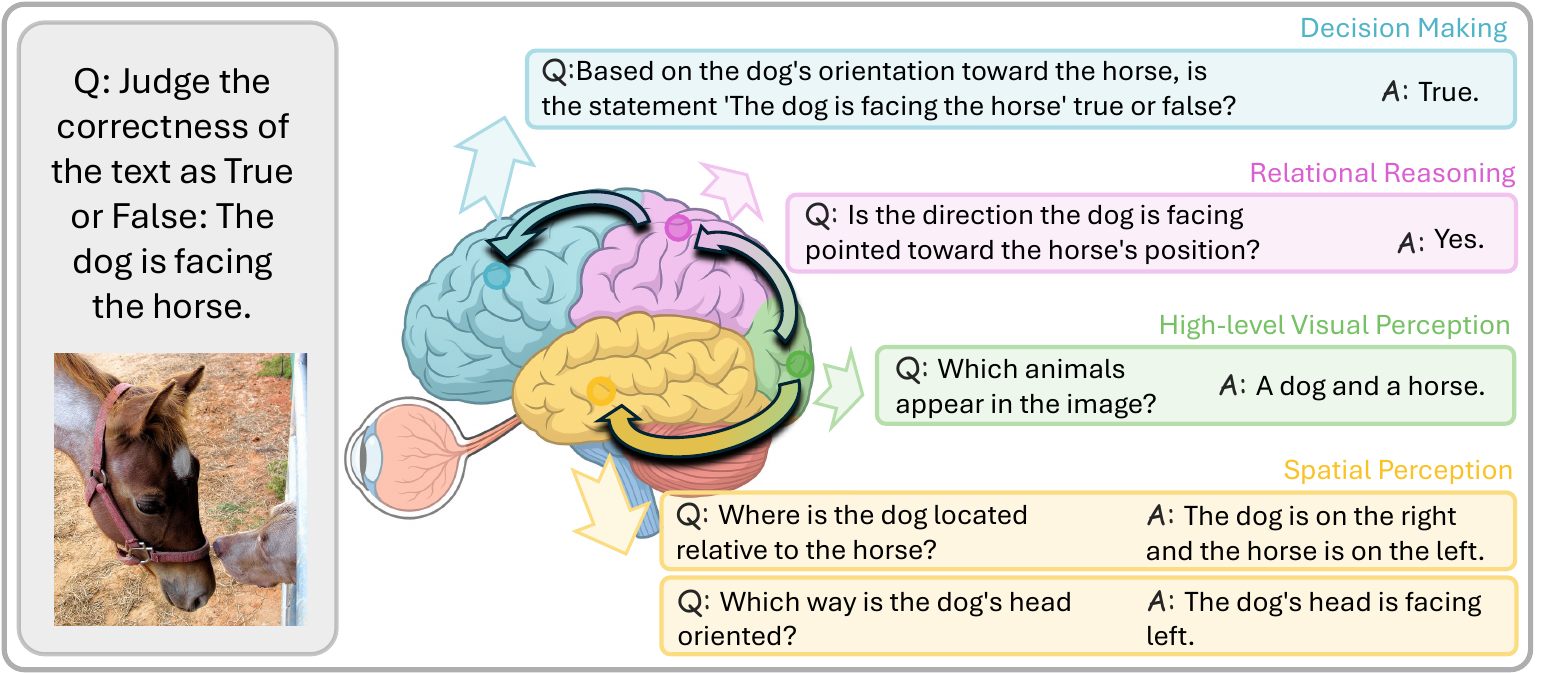} 
  \caption{To solve a complex question, the human brain engages multiple regions to perform distinct cognitive functions necessary for generating a response. We explore whether there are specific attention heads in VLMs that play functional roles in producing answers.}
  \label{fig:motivation}
\end{figure*}

Consider how humans can answer the question shown in Figure \ref{fig:motivation} by integrating perception and reasoning processes across multiple brain regions \citep{barsalou2014cognitive}: the occipital lobe, responsible for visual reception and processing the contents of the scene; the ventral pathway through the occipitotemporal lobe, supporting object recognition and high-level visual perception \cite{kravitz2013ventral}; the dorsal stream in the parietal lobe, which processes spatial relationships and enables understanding of object locations and interrelations \citep{qi2025beyond}; and the parietal and prefrontal cortices, which support higher-order relational reasoning and decision-making to produce the correct answer \citep{krawczyk2012cognition}.

To understand the \enquote{space} in VLMs, we need to examine their internal mechanisms, particularly the attention heads, which form the core of multimodal representation and alignment. Recent studies \citep{kang2025your,bi2025unveiling} have identified sparse attention heads with special functional roles for visual grounding.
However, the extent to which such specialization supports complex spatial reasoning that requires coordination across multiple functions remains largely unexplored. Bridging this gap is essential not only for interpretability but also for improving models’ spatial reasoning capabilities in a cognitively informed manner.

To facilitate this, we introduce CogVSR, a benchmark designed to disentangle spatial reasoning into interpretable cognitive components. CogVSR decomposes complex spatial reasoning questions into structured step-by-step subquestions, each corresponding to specific cognitive functions such as high-level visual perception, spatial perception, and relational reasoning, enabling fine-grained analysis of how VLMs process spatial reasoning.

Building upon CogVSR, we propose a probing framework that systematically identifies and characterizes attention heads associated with distinct cognitive functions. 
Through analysis on three major VLM families, including Intern \citep{zhu2025internvl3}, Qwen \citep{yang2025qwen3}, and Llama \citep{lin2023vila} with different model scales, our results reveal the existence of cognitive heads for different functions that consistently exhibit \textbf{universal}, \textbf{sparse}, and \textbf{intrinsic} properties across architectures. We validate the functional importance of these heads by showing that their removal degrades performance on related tasks. Further analysis shows that spatial heads, including those for spatial perception and relational reasoning, are relatively scarce and often exhibit weaker significance compared to functions such as Information Extraction and Understanding. This \textbf{scarcity of spatially specialized functions} underscores the limited spatial reasoning capacity in current VLMs. To address this, we propose a Spatial Head Activation (SHA) approach to activate latent spatial heads, yielding over 10\% accuracy improvement on InternVL3-2B for both spatial perception and relational reasoning tasks. Furthermore, we validate the importance of the identified cognitive heads on downstream spatial reasoning tasks and show that positively intervening in these functional heads can enhance the model’s spatial reasoning capabilities. In summary, our contributions are as follows:
\begin{itemize}
    \item We introduce CogVSR, a cognitively grounded benchmark for decomposing spatial reasoning into interpretable sub-processes.
    \item We develop an interpretability framework for uncovering the functional roles of attention heads in spatial reasoning.
    \item We analyze the properties of cognitive heads and develop methods to activate latent spatial heads, along with intervention strategies to enhance spatial understanding.
\end{itemize}

\section{Related Works}

\paragraph{Spatial Reasoning in VLMs.}  
Spatial reasoning, the ability to interpret and reason about object relationships in space, remains a core challenge for VLMs. Despite strong performance in image classification \cite{pratt2023does}, captioning \cite{alaluf2024myvlm}, and object detection \cite{kuo2022f}, recent studies \cite{liu2023visual, chen2025spatial, yang2025thinking} show that VLMs still struggle to capture geometric and positional relations, often failing even simple tasks such as identifying whether one object is under, above, or behind another.
To address these issues, several studies have explored specialized prompting strategies \cite{li2024topviewrs,zhang2025spatial}, post-training approaches such as supervised fine-tuning \cite{li2025llava} and reinforcement learning (RL) \cite{feng2503video}, as well as architectural modifications to VLMs \cite{ma2025spatialllm,he2025spatial,cheng2024spatialrgpt}, mainly by enhancing input representations or redesigning spatial reasoning modules. Recent studies \cite{qi2025beyond,wen2024can,chen2025spatial,zhang2025mllms} have also analyzed why MLLMs struggle with spatial reasoning, with broader discussions in the survey by \cite{zheng2025multimodal}.
While these methods have improved performance on certain spatial benchmarks and identified issues such as attention bias and insufficient geometric information, they primarily focus on simple spatial relation tasks and analyze attention at the token level. In contrast, our work provides a deeper explanation of VLMs’ behavior in complex, multi-step spatial reasoning and investigates the functional roles of internal components, i.e., attention heads, through a cognition-driven perspective.

\paragraph{Functional Roles of Attention Heads.}  
The transformer architecture, which underpins most modern VLMs, relies on multi-head attention to process and integrate information. A growing body of interpretability research has revealed that attention heads in LLMs exhibit functional specialization, supporting capabilities such as induction, truthfulness, information retrieval, and safety alignment \citep{induction,truthful,wu2404retrieval,safety,zheng2409attention}.
In the multimodal domain, recent studies \citep{li2020does} have begun to uncover the internal mechanisms of VLMs. Evidence shows that certain sparse attention heads play distinct roles in visual grounding, aligning textual tokens with image regions without additional fine-tuning \citep{kang2025your,bi2025unveiling}. Similarly, probing analyses of multimodal pre-trained models demonstrate that specific subsets of attention heads encode cross-modal interactions and semantic alignment between vision and language modalities \citep{cao2020behind}.
While these works highlight the presence of specialized heads in VLMs, they primarily focus on perception-oriented tasks such as grounding and alignment. In contrast, we investigate functionally specialized heads in the context of complex spatial reasoning.

\section{Preliminaries}

Large VLMs, such as Qwen \citep{yang2025qwen3}, typically consist of three components: a visual encoder, a language model, and a multimodal projector that connects them.
The visual encoder extracts spatially structured visual embeddings from an image, 
which are then mapped into the token space of the language model through the projector. The language model, often a transformer with $L$ stacked layers, 
contains a Multi-Head Attention (MHA) module and a feedforward network in each layer. For layer $l$, given input $X \in \mathbb{R}^{n \times d}$ 
(where $n$ is the number of tokens and $d$ the embedding dimension), 
MHA computes:
\begin{equation}
\text{MHA}^{(l)}(X) = \text{Concat}\big(N_h^{(l,1)}, \dots, N_h^{(l,H)}\big) W_o,
\end{equation}
where $H$ is the number of heads, and 
\begin{equation}
N_h^{(l,h)} = \text{Softmax}(\frac{QK^\top}{\sqrt{d_h}} + M)V.
\end{equation}
Here $Q = XW_q^{(h)}$, $K = XW_k^{(h)}$, $V = XW_v^{(h)}$, $W_q^{(h)}, W_k^{(h)}, W_v^{(h)} \in \mathbb{R}^{d \times d_h}$ are learnable projection matrices for head $h$, and $M$ denotes the attention mask matrix. 
Spatial reasoning in VLMs is supported by attention heads that integrate visual features, geometric relationships, and textual cues for inference. We analyze the functional specialization of these heads to understand their roles in spatial reasoning.

\begin{table*}[]
\caption{An example from the \textbf{CogVSR} dataset, illustrating a main question, its final answer, and a breakdown into subquestions with corresponding answers and multi-label cognitive function annotations.}
\label{tab:cogvsr-example}
\centering
\small
\setlength{\tabcolsep}{6pt}
\renewcommand{\arraystretch}{1.0}
    \resizebox{0.9\linewidth}{!}{
\begin{tabular}{p{0.2\linewidth} | p{0.3\linewidth} | p{0.15\linewidth} | p{0.2\linewidth}}
\toprule
\textbf{Main / Image} & \textbf{Subquestion} & \textbf{Answer} & \textbf{Cognitive Function Label} \\
\midrule

\multirow{5}{*}{%
\parbox{0.85\linewidth}{%
\includegraphics[width=\linewidth]{example.pdf}\\[0.5em]
\textbf{Main Question:} Judge whether the text is true or false: “The dog is facing the horse.”\\[0.3em]
\textbf{Main Answer:} True.
}%
}

& 1. Which animals appear in the image? & A dog and a horse. & High-level Visual Perception \\ \cmidrule{2-4}

& 2. Where is the dog located relative to the horse? & The dog is on the right and the horse is on the left. & \textbf{Spatial Perception, High-level Visual Perception} \\ \cmidrule{2-4}

& 3. Which way is the dog's head oriented? & The dog's head is facing left. & \textbf{Spatial Perception, High-level Visual Perception} \\ \cmidrule{2-4}

& 4. Is the direction the dog is facing pointed toward the horse's position? & Yes. & Relational Reasoning \\ \cmidrule{2-4}

& 5. Based on the dog's orientation toward the horse, is the statement “The dog is facing the horse” true or false? & True. & Decision Making \\

\bottomrule
\end{tabular}
}
\end{table*}

\section{CogVSR}

In this section, we present our Cognitive Vision Spatial Reasoning (CogVSR) dataset that contains cognitive process in spatial reasoning. CogVSR contains 1,142 main questions and 3,759 subquestions. Each example comprises a main question and answer, its subquestions and subanswers and an annotation specifying the cognitive function required for each subquestion (the example can be found in Table \ref{tab:cogvsr-example}). 

\subsection{ Cognitive Functions}

To systematically capture the cognitive processes involved in complex spatial reasoning tasks, we consider eight functions, inspired by established frameworks in cognitive science~\citep{anderson2014rules,diamond2013executive,cotabish2024spatial}.

The functions directly related to space include:
\begin{itemize}[noitemsep,topsep=0pt]
    \item \textbf{Spatial Perception}: Comprehending the spatial arrangement, positions, and geometric relationships among objects within a scene.
    
    \item \textbf{Relational Reasoning}: Understanding and comparing relationships among entities, with an emphasis on structural patterns.
    
\end{itemize}

Other functions related to spatial reasoning include:
\begin{itemize}[noitemsep,topsep=0pt]
\item \textbf{Low-level Visual Perception}: Recognizing basic visual features such as color, shape, size. 

\item \textbf{High-level Visual Perception}: Integrating visual information to recognize objects, patterns, and scene structure at a semantic level. 

\item \textbf{Language Information Extraction and Understanding}: locating and understanding relevant information from an external source or prior context. 

\item \textbf{Knowledge Recall}: Accessing long-term knowledge including language and vision. 

\item \textbf{Math Reasoning}: Performing counting, arithmetic, comparison, and logic-based operations. 

\item \textbf{Decision-Making}: Selecting the best outcome among alternatives based on reasoning. 

\end{itemize}

This categorization reflects a natural progression from basic information processing to complex cognitive integration. Both the human brain and VLMs encompass a wide range of functional modules. Our focus in this work is specifically on spatial reasoning-related  cognitive functions. By identifying and organizing these eight core functions, we can more clearly examine how VLMs handle different types of thinking steps, in a way that is both systematic and easy to interpret. The detail descriptions of each function and more examples in CogVSR can be found in the Appendix.

\subsection{Data Collections}
Based on our categorization of reasoning functions, we sampled 1600 diverse questions from existing commonly used visual reasoning benchmarks, selecting 400 examples each from VSR \cite{Liu2022VisualSR}, SpatialEval \cite{wang2024picture}, 3DSRbench \cite{ma20253dsrbench},  Spatial457 \cite{wang2025spatial457} dataset. These datasets cover a broad spectrum of spatial reasoning tasks, ranging from simple position and orientation judgments to more complex challenges such as occlusion, counting, path planning, and maze navigation. Using the chain-of-thought (CoT) paradigm, we prompted GPT-o4-mini~\citep{openai_o3_o4mini_system_card_2025} to decompose each main question (with its final answer) into subquestions, each targeting a subanswer and a specific function. The prompt encourages structured, step-by-step reasoning, ensuring that each subquestion is clear, answerable, and sequentially dependent. This process yields a set of subquestion–answer–function (subQAF) triples for each QA pair: $\operatorname{subQAFs}=\left\{\left(q_i, a_i, f_i\right)\right\}_{i=1}^k$, where each contains a subquestion $q_i$, its concise subanswer $a_i$, and the corresponding function label $f_i$. The prompt for generating subquestions are list in Appendix.

\subsection{Data Filtering and Annotation}

Recent advances enable leveraging large pre-trained models for dataset construction, thanks to their reasoning capabilities and ability to generate high-quality annotations at scale~\citep{llm_annotate}. While our dataset is automatically generated using a large VLM to minimize manual effort, we employ a rigorous two-stage human verification pipeline to ensure data quality and mitigate hallucination. In the first stage, three expert annotators
independently evaluate whether each subquestion is logically structured and consistent with natural human reasoning, as well as whether each assigned cognitive function is accurate. QA pairs with incoherent or inconsistent decompositions are filtered out. In the second stage, for the retained subquestions, annotators relabel the cognitive functions where disagreement occurred, ensuring accurate alignment with the intended mental processes. Notably, although GPT-o4-mini generates one primary function for each subquestion, a single subquestion may actually involve multiple cognitive functions, for example, spatial perception (e.g., orientation) accompanied by high-level visual perception (e.g., object recognition), \textcolor{black}{as illustrated by subquestion 2 in Table \ref{tab:cogvsr-example}.} Therefore, we adopt a multi-label annotation approach. Simultaneously, subanswers are reviewed and, if necessary, corrected by human annotators (details in Appendix).

This multi-step verification ensures that each retained subQAF triplet represents a coherent, interpretable reasoning step grounded in core cognitive functions. After this refinement, our final dataset comprises 1,142 main QA pairs and 3,759 validated subQAF triplets. Detailed dataset statistics, including training and testing triplets and function distributions, are provided in the Appendix.

\begin{figure*}[]
    \centering
    \includegraphics[width=0.88\textwidth]{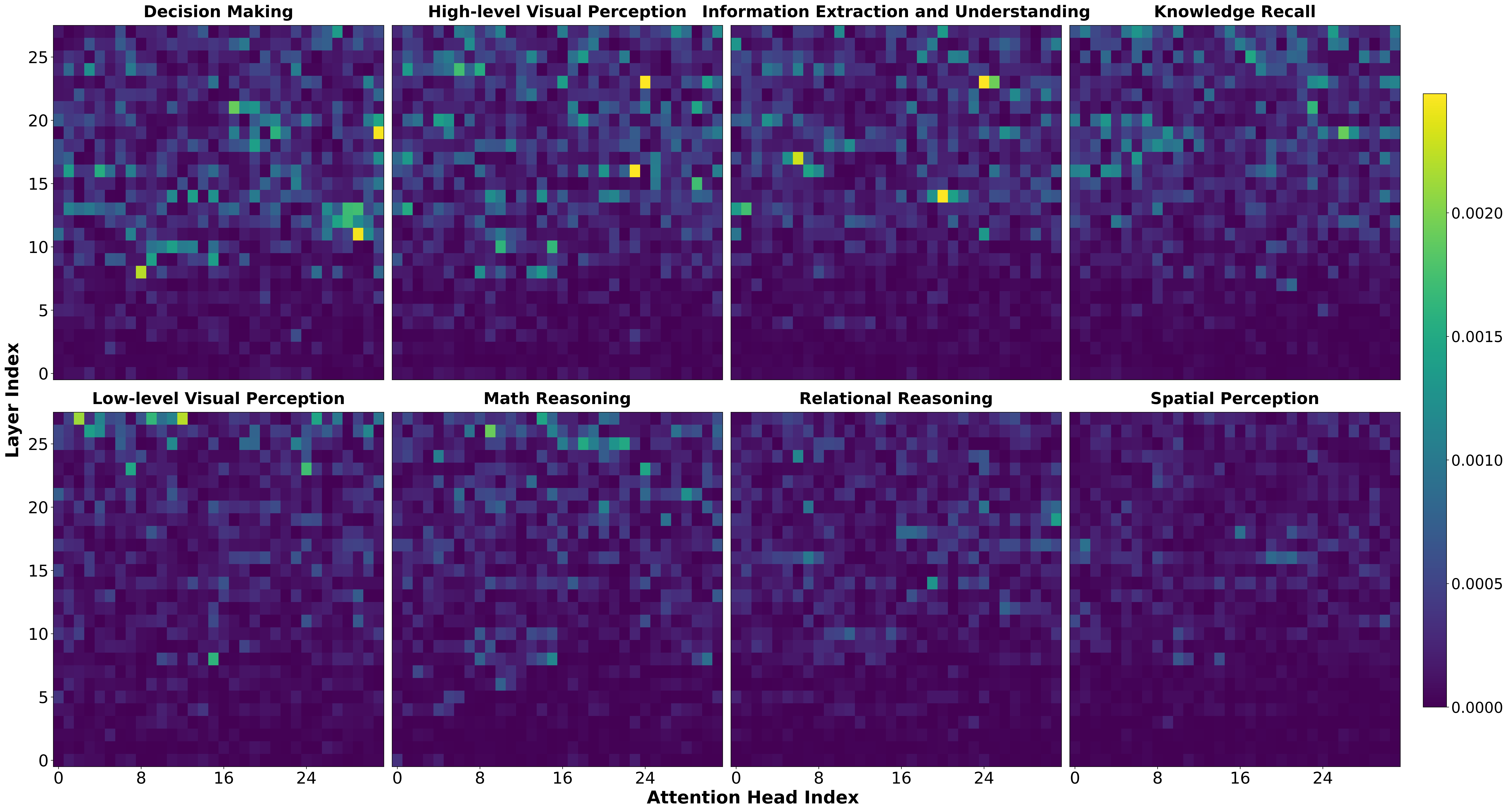}
    \caption{The existence of cognitive heads in Qwen2.5-VL-7B responsible for eight distinct functions in complex reasoning tasks. The x-axis represents the head index, while the y-axis indicates the layer index. The values denote head importance scores.}
    \label{fig:heatmap}
\end{figure*}

\section{Attention Heads Related to Spatial Reasoning}

Using the CogVSR dataset, we adopt a probing-based framework \citep{alain2016understanding, belinkov2022probing, tenney2019bert} to identify which attention heads in VLMs are associated with specific functions in the reasoning process. Specifically, for each functionally annotated subquestion, we extract head activations (see Subsection~\ref{extract}), train a multi-label classifier and compute importance scores to identify contributing heads (see Subsection~\ref{probing}). 

\subsection{Head Feature Extraction}
\label{extract}

Given a large VLM $\mathcal{M}$, we generate an answer $a_i^{\mathcal{M}}$ for each subquestion $q_i$ derived from a main question $Q_i$. To support coherent multi-step reasoning, we include preceding subquestions and their correct answers as contextual input, emulating the incremental reasoning process observed in human cognition.

During inference, each input token is first mapped into an embedding and then propagated through the transformer’s layers. At each layer, attention and feedforward operations update the residual stream, which is ultimately decoded into token predictions. For each generated token $i$, we extract attention head outputs $X_i = \{ x_l^m(i) \mid l = 1, \dots, L,\ m = 1, \dots, M \}$ across all layers, where $x^m_l$ denotes the value vector from the $m$-th head in layer $l$ projected into the residual stream, with $M$ the number of heads per layer and $L$ the total number of layers.

Let $N_t$ denote the number of tokens in the generated answer $a_i^{\mathcal{M}}$. To isolate semantically informative content relevant to reasoning, we select the top-$k$ most important tokens, determined by prompting Gemini2.5-Flash \cite{comanici2025gemini}, yielding an index set $\mathcal{I}_k$ with $|\mathcal{I}_k| = k$. For each index $j \in \mathcal{I}_k$, we extract the corresponding attention head activations from $X_j$, and compute the averaged activation feature for the $m$-th head in layer $l$ as $\bar{x}_l^m = \frac{1}{k} \sum_{j \in \mathcal{I}_k} x_l^m(j)$. Given prior findings suggesting that cognitive functions may vary by layer depth~\cite{zheng2409attention}, we incorporate layer-wise information by computing the average activation $\bar{x}_l=\frac{1}{M} \sum_{m=1}^M \bar{x}_l^m$ for each layer. We then augment each head-level vector with its corresponding layer summary, resulting in enriched features $\bar{x}^{m'}_l = [\bar{x}^m_l;\bar{x}_l]$. 


\subsection{Function Probing}
\label{probing}

For the dataset with $N$ subQAF triplets, we collect all activations to construct the probing dataset:

\begin{equation}
    \mathcal{D}_{\text{probe}} = \left\{ (\bar{x}^{m'}_l,\ c)_i \right\}_{i=1}^{N}, l \in \{1, \ldots, L\},\ m \in \{1, \ldots, M\}
\end{equation}


For classification based on CogVSR, the training set includes 909 main questions with 2,969 subQAF triplets, while the testing set has 233 main questions with 790 triplet. Notably, we use the attention-head output values only when the VLM $\mathcal{M}$ generates the correct answer for the subquestions. Each attention head feature is first passed through a trainable linear projection for dimensionality reduction, after which all projected features are concatenated into a single vector and fed into a two-layer MLP for multi-label classification over cognitive functions (training details are provided in Appendix). To interpret the contribution of individual heads to each function, we use a gradient-based attribution method. Specifically, for each function class \(c\), we compute the contribution of each head feature via the gradient$\times$activation technique:

\begin{equation}
    I^{(c)}_j = \mathbb{E}_{(\bar{x}, c) \sim \mathcal{D}_{\text{probe}}} \left[ \frac{\partial \hat{y}_c}{\partial \bar{x}_j} \cdot \bar{x}_j \right],
\end{equation}
where \(\bar{x}_j\) is the \(j\)-th head input feature, and \(\hat{y}_c\) is the classifier’s predicted logit for class \(c\). This yields an importance score for each attention head with respect to each cognitive function. We aggregate the scores into a matrix \(\mathbf{I} \in \mathbb{R}^{C \times (L \cdot M)}\), where each row corresponds to a function class and each column to a specific head in a specific layer.

We hypothesize that attention heads with higher importance scores contribute more significantly to each cognitive function. By ranking heads according to their importance, we can identify which heads and layers are specialized for specific functions. Subsequent targeted interventions on these heads validate the effectiveness of this approach.
The effectiveness of top-k ($k=3$) tokens and layer information can be found in Appendix.

\begin{table*}[]
    \centering
    \setlength{\tabcolsep}{3pt}
        \caption{Intervention results (\%) of cognitive heads vs. random heads across 8 functions: \textbf{Spatial} Perception, \textbf{Relational} Reasoning, \textbf{Low-level} Visual Perception, \textbf{High-level} Visual Perception, Language \textbf{Info}rmation Extraction and Understanding, Knowledge \textbf{Recall}, \textbf{Math} Reasoning, and \textbf{Decision} Making. For model names, Intern2B: InternVL3-2B, Intern8B: InternVL3-8B, Llama11B: Llama3.2-11B-Vision, Llama90B:  Llama3.2-90B-Vision, Qwen3B: Qwen2.5-VL-3B-Instruct, Qwen7B: Qwen2.5-VL-7B-Instruct. Lower values indicate more effective intervention outcomes, suggesting that the corresponding heads play a greater role in the cognitive function.}
    \resizebox{1\linewidth}{!}{
    \renewcommand{\arraystretch}{1.0}
    \begin{tabular}{lc|cccccccccccccccc}
        \toprule
        \multirow{3}{*}{Model} & \multirow{3}{*}{Inter\_Head} 
        & \multicolumn{4}{c}{Spatial Functions} 
        & \multicolumn{12}{c}{Other Functions} \\
        \cmidrule(r){3-6} \cmidrule(r){7-18}
        & & \multicolumn{2}{c}{Spatial} & \multicolumn{2}{c}{Relational} &  \multicolumn{2}{c}{Low-Level} & \multicolumn{2}{c}{High-Level}
        & \multicolumn{2}{c}{Info} & \multicolumn{2}{c}{Recall}
        & \multicolumn{2}{c}{Math} & \multicolumn{2}{c}{Decision} \\
        \cmidrule(r){3-4} \cmidrule(r){5-6} \cmidrule(r){7-8} \cmidrule(r){9-10}
        \cmidrule(r){11-12} \cmidrule(r){13-14} \cmidrule(r){15-16} \cmidrule(r){17-18}
        & & llm & acc & llm & acc & llm & acc & llm & acc 
        & llm & acc & llm & acc & llm & acc & llm & acc \\
        \midrule
        \multirow{2}{*}{Intern2B} 
        & random    & 61.37 & 56.82 & 63.25 & 58.13 & 65.48 & 60.91 & 69.72 & 64.35 & 71.19 & 66.04 & 69.87 & 64.11 & 59.44 & 54.27 & 64.03 & 58.76 \\
        & cognitive &  \textbf{37.92} &  \textbf{34.37} & \textbf{31.56} &  \textbf{26.88} & \textbf{13.43} &  \textbf{18.21} & \textbf{34.27} & \textbf{28.65} & \textbf{39.82} & \textbf{33.17} & \textbf{37.11} & \textbf{31.49} & \textbf{23.68} & \textbf{18.02} & \textbf{21.54} & \textbf{16.43} \\
        \midrule
        \multirow{2}{*}{Intern8B} 
        & random    & 72.84 & 68.12 & 74.33 & 69.57 & 76.05 & 71.44 & 81.26 & 76.09 & 83.17 & 78.03 & 81.92 & 76.48 & 73.66 & 68.40 & 77.41 & 72.05 \\
        & cognitive & \textbf{46.39} & \textbf{42.07} & \textbf{52.84} & \textbf{48.51} & \textbf{36.12} & \textbf{31.76} & \textbf{59.73} & \textbf{53.28} & \textbf{65.04} & \textbf{58.69} & \textbf{62.91} & \textbf{56.33} & \textbf{49.82} & \textbf{44.15} & \textbf{46.37} & \textbf{40.88} \\
        \midrule
        \multirow{2}{*}{Llama11B} 
        & random    & 74.95 & 70.44 & 76.68 & 72.11 & 78.37 & 73.90 & 83.29 & 78.16 & 85.06 & 79.87 & 83.64 & 78.33 & 75.92 & 70.83 & 79.18 & 74.06 \\
        & cognitive & \textbf{44.73} & \textbf{40.58} & \textbf{50.41} & \textbf{56.33} & \textbf{34.86} & \textbf{30.17} & \textbf{57.24} & \textbf{51.63} & \textbf{62.77} & \textbf{56.94} & \textbf{60.58} & \textbf{54.72} & \textbf{47.39} & \textbf{41.92} & \textbf{43.85} & \textbf{38.14} \\
        \midrule
        \multirow{2}{*}{Llama90B} 
        & random    & 86.21 & 82.07 & 88.03 & 83.92 & 89.77 & 85.63 & 93.18 & 89.02 & 94.56 & 90.41 & 93.02 & 88.93 & 87.49 & 83.36 & 91.07 & 86.95 \\
        & cognitive & \textbf{72.44} & \textbf{68.26} & \textbf{76.71} & \textbf{72.39} & \textbf{60.03} & \textbf{55.89} & \textbf{74.82} & \textbf{69.44} & \textbf{79.15} & \textbf{73.86} & \textbf{77.03} & \textbf{71.74} & \textbf{69.42} & \textbf{64.13} & \textbf{66.85} & \textbf{61.27} \\
        \midrule
        \multirow{2}{*}{Qwen3B} 
        & random    & 69.12 & 64.88 & 71.03 & 66.74 & 73.49 & 69.05 & 77.85 & 73.12 & 79.64 & 75.02 & 78.03 & 73.41 & 69.77 & 65.23 & 73.54 & 69.01 \\
        & cognitive &  \textbf{35.63} &  \textbf{42.94} & \textbf{31.28} &  \textbf{47.46} & \textbf{14.92} & \textbf{10.08} & \textbf{38.47} & \textbf{32.80} & \textbf{44.35} & \textbf{38.06} & \textbf{41.72} & \textbf{35.94} & \textbf{27.58} & \textbf{22.16} & \textbf{25.03} & \textbf{19.77} \\
        \midrule
        \multirow{2}{*}{Qwen7B} 
        & random    & 80.04 & 75.92 & 82.11 & 77.86 & 84.39 & 80.21 & 88.27 & 84.10 & 90.02 & 85.73 & 88.39 & 84.03 & 82.76 & 78.44 & 86.51 & 82.19 \\
        & cognitive & \textbf{57.69} & \textbf{53.42} & \textbf{63.85} & \textbf{59.57} & \textbf{47.92} & \textbf{43.18} & \textbf{68.31} & \textbf{63.04} & \textbf{73.77} & \textbf{68.55} & \textbf{71.23} & \textbf{65.98} & \textbf{62.84} & \textbf{57.63} & \textbf{59.41} & \textbf{54.22} \\
        \bottomrule
    \end{tabular}
    }
    \label{tab:vlm_results}
\end{table*}

\section{Experiments}

We conduct a series of experiments on three VLM families across various model scales, including Intern \citep{zhu2025internvl3} (InternVL3-2B and InternVL3-8B), Qwen \citep{yang2025qwen3} (Qwen2.5-VL-7B and Qwen2.5-VL-3B), and Llama3.2-Vision \citep{liu2025feasibility} (Llama-3.2-11B-Vision and Llama-3.2-90B-Vision). We analyze the common properties of functional heads (Subsection~\ref{properties}) and validate their contributions (Subsection~\ref{contribution}). We further investigate the scarcity of spatial heads (Subsection~\ref{scarce}), which likely contributes to the failure of VLMs in spatial reasoning. We propose methods to activate latent spatial heads, thereby enhancing performance. Additionally, we assess the causal impact of cognitive heads on downstream reasoning tasks (Subsection~\ref{tasks}); our results show that modulating spatially related heads can improve spatial reasoning.

\subsection{Properties of Cognitive Heads}
\label{properties}


\textbf{Sparsity, Universality, and Intrinsic Organization:} Figure~\ref{fig:heatmap} shows the heatmap of head importance scores across eight functions in Qwen2.5-VL-7B on the CogVSR test set, revealing a sparse distribution. In total, fewer than 9\% of all heads have importance scores above 0.001 across the eight functions (about 3\% for high-level visual perception and decision making, and less than 1\% for the others),
suggesting that only a compact subset of heads meaningfully contribute to task performance. These results demonstrate that VLMs rely on highly specialized, localized components for distinct  cognitive abilities. Moreover, this sparse functional organization is consistent across architectures and scales: heatmaps for five additional models (in Appendix) confirm its universality. Within the same model family (e.g., Qwen2.5-VL-7B vs. Qwen2.5-VL-3B), we observe similar distributions, suggesting that such specialization is intrinsic to VLMs.

\subsection{Functional Contributions of Cognitive Heads}
\label{contribution}
We define the heads with high importance scores for each function as cognitive heads.
Specifically, we identify cognitive heads by selecting those before the elbow point of each function’s descending importance curve (see Appendix), and find notable variation in head counts across functions.

After identifying the cognitive heads associated with each function, we examine their functional roles by evaluating the model's behavior on the CogVSR test set under targeted interventions.
We perform head ablation by scaling the output of a specific attention head with a small factor $\epsilon$ (e.g., 0.001), effectively suppressing its contribution:

\begin{equation}
x_i^{\text{mask}} = \operatorname{Softmax}\left(\frac{W_q^i W_k^{i T}}{\sqrt{d_k / n}}\right) \cdot \epsilon W_v^i
\end{equation}
Specifically, we compare model performance when masking identified cognitive heads versus masking an equal number of randomly-selected heads. To quantify the impact, we employ both an LLM-based judge (Gemini2.5-flash-lite \cite{comanici2025gemini}) and an integrated accuracy metric. An output is considered unaffected if its BLEU score~\citep{papineni2002bleu} exceeds 0.8, or if either the ROUGE score~\citep{chin2004rouge} or the semantic similarity score surpasses 0.6. This provides a comprehensive evaluation of performance degradation.


As shown in Table~\ref{tab:vlm_results}, masking cognitive heads leads to a substantial decline in performance, whereas masking an equal number of random heads results in only minor degradation across all VLMs. In some cases, masking the identified cognitive heads reduces accuracy to below 20\%, indicating that the model can almost no longer perform the corresponding function without them.


\begin{figure}[H]
    \centering
    \includegraphics[width=0.9\linewidth]{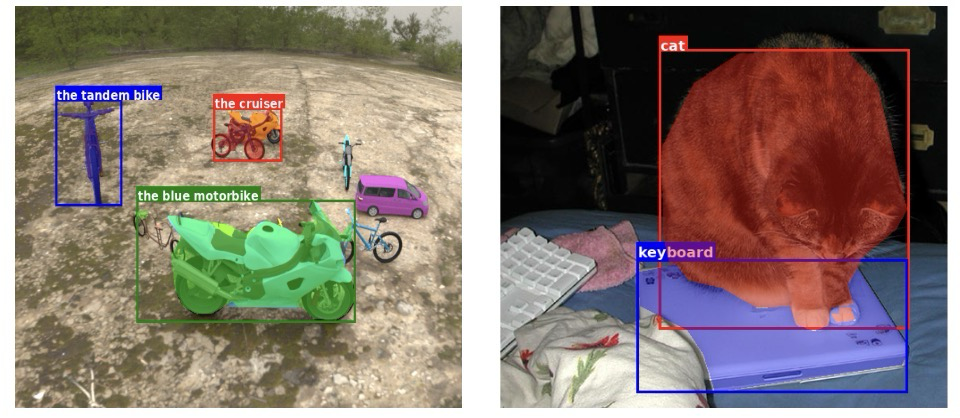}
    \caption{Objects with applied masks and bounding boxes as input.}
    \label{fig:bbx}
\end{figure}
\subsection{Scarcity of Spatially-Related Heads}
\label{scarce}
By closely examining spatially related attention heads, we find that, in the context of the generally sparse cognitive heads in VLMs, spatial and relational reasoning heads are noticeably scarcer than those associated with other functions, with relatively few exhibiting significant importance scores. These spatial and relational reasoning heads are primarily responsible for encoding object positions, orientations, geometric layouts, and object-to-object relations, playing a particularly important role in spatial reasoning tasks. Their scarcity directly impacts performance on such tasks, suggesting that current VLMs likely underrepresent spatial reasoning due to this imbalance.

\subsection{Active Latent Spatial Heads}
\begin{figure*}
    \centering
    \includegraphics[width=0.88\textwidth]{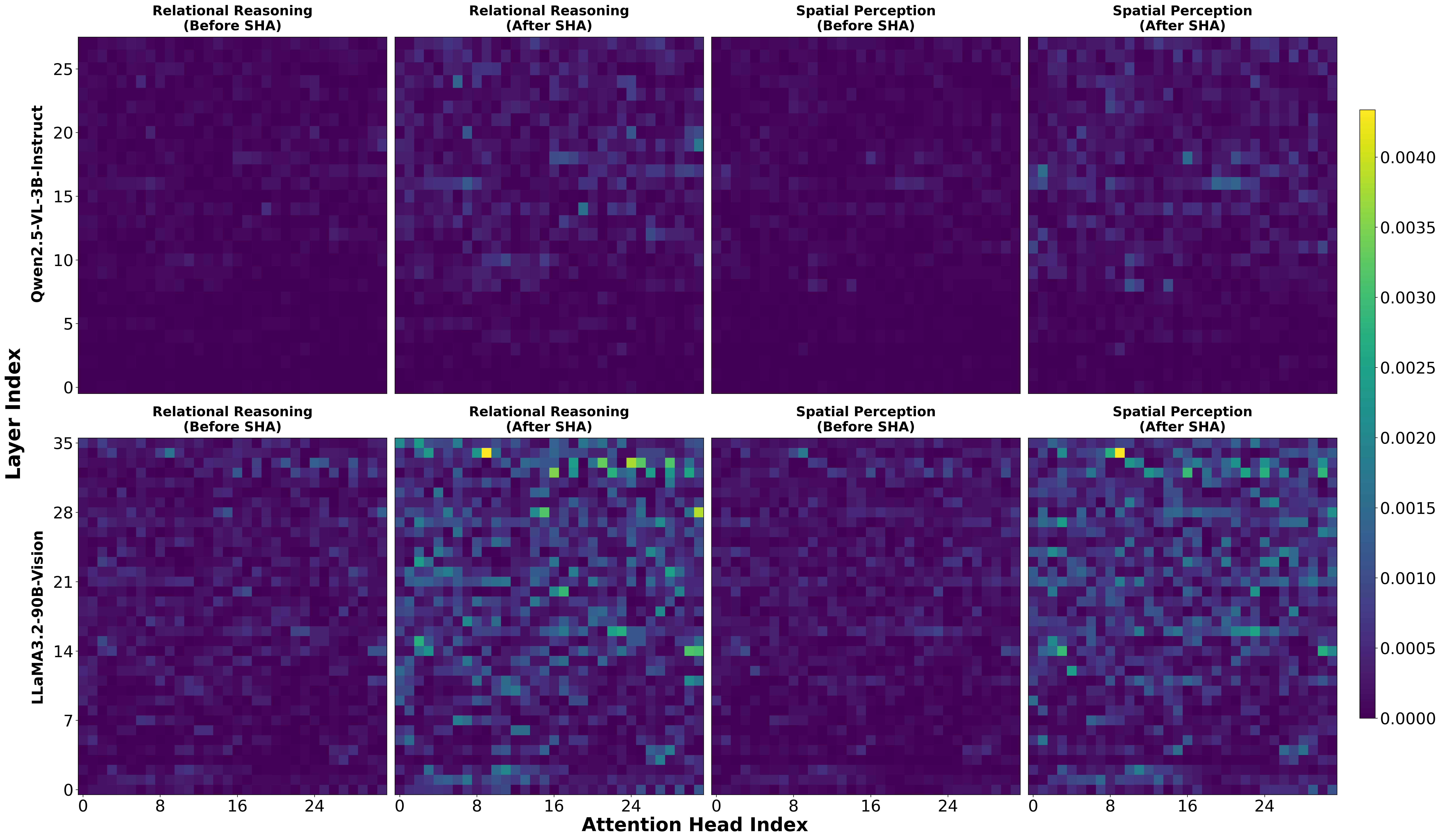}
    \caption{Comparison of spatial perception and relational reasoning heads before and after spatial head activation across various models. }
    \label{fig:activation}
\end{figure*}

How can we encourage attention heads in VLMs to focus on spatially related functions, i.e., to activate latent spatial heads? We propose a Spatial Head Activation (SHA) approach that disentangles high-level visual perception (e.g., object recognition) from spatial perception. Concretely, we use Gemini2.5-Flash~\cite{balestri2025gender} to detect objects and obtain bounding boxes, and then apply masking to the detected regions to provide additional object priors while reducing the model’s reliance on high-level visual cues (see Figure \ref{fig:bbx}). This design encourages the model, when answering spatial perception and relational reasoning subquestions, to more effectively engage spatially oriented capabilities and activate latent spatial heads. As illustrated in Figure~\ref{fig:activation}, after applying SHA, both Qwen2.5-VL-3B and Llama3.2-90B-Vision exhibit an increased number of spatial and relational reasoning heads. Consistently, Table \ref{tab:augmentation_spatial} shows that InternVL3-2B and Llama3.2-90B-Vision achieve about 10\% and 5\% accuracy improvements, respectively, on both spatial perception and relational reasoning tasks when applying SHA.
\subsection{Influence of Functional Heads on Downstream Tasks}
\label{tasks}
In this section, we investigate how functional heads influence downstream tasks through both negative interventions (masking out function heads) and positive interventions (shifting heads toward specific functions).

\begin{table}[]
    \setlength{\tabcolsep}{6pt}
    \caption{Effect of spatial input augmentation on spatial and relational functions. \enquote{Original} uses only the raw image; \enquote{+BBox} and \enquote{+Mask} add bounding-box and instance-mask cues respectively; \enquote{+BBox+Mask (Ours)} incorporates both.}
    \resizebox{\linewidth}{!}{%
        \renewcommand{\arraystretch}{1}
        \begin{tabular}{lc|cccc}
            \toprule
            \multirow{2}{*}{Model} & \multirow{2}{*}{Input} 
            & \multicolumn{2}{c}{Spatial} 
            & \multicolumn{2}{c}{Relational} \\
            \cmidrule(r){3-4} \cmidrule(r){5-6}
            & & llm & acc & llm & acc \\
            \midrule
            \multirow{4}{*}{Intern2B}
            & Original          & 61.37 & 56.82 & 63.25 & 58.13 \\
            & +BBox             & 65.84 & 58.94 & 64.95 & 62.39 \\
            & +Mask             & 59.70 & 59.80 & 66.82 & 56.47 \\
            & \textbf{+BBox+Mask (Ours)} 
                                & \textbf{71.82} & \textbf{68.64} & \textbf{70.12} & \textbf{67.04} \\
            \midrule
            \multirow{4}{*}{Llama90B}
            & Original          & 86.21 & 82.07 & 88.03 & 83.92 \\
            & +BBox             & 90.07 & 86.93 & 89.84 & 85.75 \\
            & +Mask             & 87.57 & 83.38 & 91.22 & 85.11 \\
            & \textbf{+BBox+Mask (Ours)} 
                                & \textbf{92.90} & \textbf{87.77} & \textbf{92.41} & \textbf{89.27} \\
            \bottomrule
        \end{tabular}
    }%
    \label{tab:augmentation_spatial}
\end{table}

\begin{table}[]
    \setlength{\tabcolsep}{6pt}
    \caption{Negative intervention on visual spatial reasoning benchmarks. Scores are LLM-Judge accuracy (\%). ``before'' uses the original model; ``after'' applies intervention.}
    \renewcommand{\arraystretch}{1.0}
        \resizebox{\linewidth}{!}{
    \begin{tabular}{l c | ccc}
        \toprule
        {Dataset} & {Inter\_Head} & 
        Intern8B & Qwen7B & Llama11B \\
        \midrule
        \multirow{2}{*}{VSR}    & before & 64.35 & 62.47 & 65.82 \\
                     & after  & \textbf{24.13} & \textbf{21.76} & \textbf{17.94} \\
        \midrule
        \multirow{2}{*}{Spatial457}   & before & 41.23 & 38.71 & 32.18 \\
                     & after  & \textbf{14.97} & \textbf{12.43} & \textbf{9.86}  \\
        \midrule
        \multirow{2}{*}{SpatialEval}  & before & 47.92 & 45.33 & 39.61 \\
                     & after  & \textbf{20.08} & \textbf{18.57} & \textbf{15.19} \\
        \midrule
        \multirow{2}{*}{3DSRBench}    & before & 36.58 & 34.16 & 28.87 \\
                     & after  & \textbf{12.39} & \textbf{10.84} & \textbf{8.72}  \\
        \bottomrule
    \end{tabular}
    }
    \label{tab:negative_intervention_spatial}
\end{table}

\begin{table*}[]
    \centering
    \setlength{\tabcolsep}{3pt}
    \caption{Positive intervention on spatial reasoning benchmarks (200 samples per dataset). Scores are LLM-Judge accuracy (\%). For out-of-domain data, we positively intervene on spatial perception and relational reasoning heads.}
    \resizebox{0.85\linewidth}{!}{%
    \renewcommand{\arraystretch}{1.0}
    \begin{tabular}{lc|cccccccc|cccc}
        \toprule
        \multirow{2}{*}{Model} & \multirow{2}{*}{Inter\_Head} 
        & \multicolumn{8}{c|}{In Domain} 
        & \multicolumn{4}{c}{Out of Domain} \\
        \cmidrule(r){3-10} \cmidrule(r){11-14}
        & & Spatial & Relational & Low & High & Info & Recall & Math & Decision
          & VSR & Spatial457 & SpatialEval & 3DSRBench \\
        \midrule
        \multirow{2}{*}{Qwen7B} 
        & before & 80.04 & 82.11 & \textbf{84.39} & 88.27 & 90.02 & 88.39 & 82.76 & 86.51
                 & 62.47 & 38.71 & 45.33 & 34.16 \\
        & after  & \textbf{81.89} & \textbf{83.76} & 84.10 & \textbf{90.63} 
                 & \textbf{92.41} & \textbf{90.92} & \textbf{82.90} & \textbf{87.74}
                 & \textbf{63.72} & \textbf{39.15} & \textbf{45.81} & \textbf{34.95} \\
        \midrule
        \multirow{2}{*}{Llama11B} 
        & before & 74.95 & 76.68 & 78.37 & 83.29 & 85.06 & 83.64 & 75.92 & 79.18
                 & 65.82 & 32.18 & 39.61 & 28.87 \\
        & after  & \textbf{75.62} & \textbf{77.40} & \textbf{79.10} & \textbf{83.75}
                 & \textbf{86.20} & \textbf{83.10} & \textbf{75.40} & \textbf{79.65}
                 & \textbf{66.91} & \textbf{32.40} & \textbf{39.90} & \textbf{29.40} \\
        \midrule
        \multirow{2}{*}{Intern8B} 
        & before & 72.84 & 74.33 & 76.05 & 81.26 & 83.17 & 81.92 & \textbf{73.66} & 77.41
                 & 64.35 & 41.23 & 47.92 & 36.58 \\
        & after  & \textbf{74.55} & \textbf{74.90} & \textbf{77.80} & \textbf{81.90}
                 & \textbf{83.50} & \textbf{82.60} & 73.20 & \textbf{79.89}
                 & \textbf{65.28} & \textbf{41.85} & \textbf{48.45} & \textbf{37.20} \\
        \bottomrule
    \end{tabular}
    }%
    \label{tab:positive_intervention_spatial_llm}
\end{table*}

\paragraph{Negative Intervention:}
We randomly sample 200 question for four benchmarks, VSR \cite{Liu2022VisualSR}, Spatial457 \cite{wang2025spatial457}, SpatialEval \cite{wang2024picture}, and 3DSRBench \cite{ma20253dsrbench}, both spatial reasoning tasks. We perform negative intervention by masking spatial perception heads of VLMs, effectively suppressing their activations. 
As shown in Table \ref{tab:negative_intervention_spatial}, masking these key cognitive heads leads to a significant performance drop across different models and datasets. 

\paragraph{Positive Intervention:} We calculate the activation directions of different functions using the CogVSR dataset. For each function, the activation direction of a head at layer $l$ and index $h$ is computed as:
\begin{equation}
\operatorname{dir}_l^h=\mathbb{E}_{i \in \mathcal{D}_{\text {correct }}}\left[x_l^h(i)\right]-\mathbb{E}_{i \in \mathcal{D}_{\text {incorrect }}}\left[x_l^h(i)\right]
\end{equation}
where $x_l^h(i)$ denotes the activation of head at layer $l$ and index $h$, and $\mathcal{D}_{\text {correct }}$ and $\mathcal{D}_{\text {incorrect }}$ represent the sets of samples answered correctly and incorrectly, respectively. Then we estimate the standard deviation of activations \citep{truthful} along the cognitive function direction to be $\sigma_l^h$, and shift original head activation as $x_l^h(i) \leftarrow x_l^h(i)+\alpha \sigma_l^h \operatorname{dir}_l^h$, where $\alpha$ is a parameter. 

The experimental results in Table \ref{tab:positive_intervention_spatial_llm} show that enhancing the activation of functional heads along their corresponding functional directions improves performance on the related tasks. For example, positive intervention on spatial perception heads in Qwen2.5-VL-7B increased accuracy on the corresponding CogVSR spatial perception subquestions from 80.04\% to 81.89\%. Similarly, enhancing spatial related heads can also boost performance on downstream tasks. Here, we set $\alpha = 0.1$ for all datasets, though tuning this parameter may further improve performance.

\section{Conclusion}
In this work, we conducted a mechanistic investigation into how VLMs perform spatial reasoning through the lens of attention head specialization. By introducing CogVSR, a cognitively grounded benchmark that decomposes spatial reasoning into interpretable subfunctions such as spatial perception and relational reasoning, we systematically identified attention heads responsible for these cognitive processes.
Across multiple VLM families, we find a consistent scarcity of spatially specialized heads, suggesting underrepresentation of spatial understanding. Targeted activation of latent spatial heads improves spatial awareness without retraining, and intervention experiments confirm their critical role: ablating them harms performance, while emphasizing them enhances accuracy.
Our findings shed light on the internal organization of VLMs and emphasize the need for more interpretable and cognitively inspired VLMs. However, several limitations remain: we focus on eight predefined cognitive functions and restrict our analysis to attention heads, leaving other components such as MLPs unexplored. Future work may employ advanced probing methods and extend the analysis to offer a deeper understanding of spatial reasoning in VLMs.


{
    \small
    \bibliographystyle{ieeenat_fullname}
    \bibliography{main}

@String(ICCV= {Int. Conf. Comput. Vis.})

@String(ICCV  = {ICCV})

@inproceedings{yang2025thinking,
  title={Thinking in space: How multimodal large language models see, remember, and recall spaces},
  author={Yang, Jihan and Yang, Shusheng and Gupta, Anjali W and Han, Rilyn and Fei-Fei, Li and Xie, Saining},
  booktitle={Proceedings of the Computer Vision and Pattern Recognition Conference},
  pages={10632--10643},
  year={2025}
}

@article{zhu2023minigpt,
  title={Minigpt-4: Enhancing vision-language understanding with advanced large language models},
  author={Zhu, Deyao and Chen, Jun and Shen, Xiaoqian and Li, Xiang and Elhoseiny, Mohamed},
  journal={arXiv preprint arXiv:2304.10592},
  year={2023}
}

@article{liu2023visual,
  title={Visual instruction tuning},
  author={Liu, Haotian and Li, Chunyuan and Wu, Qingyang and Lee, Yong Jae},
  journal={Advances in neural information processing systems},
  volume={36},
  pages={34892--34916},
  year={2023}
}

@article{lu2024deepseek,
  title={Deepseek-vl: towards real-world vision-language understanding},
  author={Lu, Haoyu and Liu, Wen and Zhang, Bo and Wang, Bingxuan and Dong, Kai and Liu, Bo and Sun, Jingxiang and Ren, Tongzheng and Li, Zhuoshu and Yang, Hao and others},
  journal={arXiv preprint arXiv:2403.05525},
  year={2024}
}

@article{qi2025beyond,
  title={Beyond semantics: Rediscovering spatial awareness in vision-language models},
  author={Qi, Jianing and Liu, Jiawei and Tang, Hao and Zhu, Zhigang},
  journal={arXiv preprint arXiv:2503.17349},
  year={2025}
}

@book{barsalou2014cognitive,
  title={Cognitive psychology: An overview for cognitive scientists},
  author={Barsalou, Lawrence W},
  year={2014},
  publisher={Psychology Press}
}

@article{krawczyk2012cognition,
  title={The cognition and neuroscience of relational reasoning},
  author={Krawczyk, Daniel C},
  journal={Brain research},
  volume={1428},
  pages={13--23},
  year={2012},
  publisher={Elsevier}
}

@inproceedings{kang2025your,
  title={Your large vision-language model only needs a few attention heads for visual grounding},
  author={Kang, Seil and Kim, Jinyeong and Kim, Junhyeok and Hwang, Seong Jae},
  booktitle={Proceedings of the Computer Vision and Pattern Recognition Conference},
  pages={9339--9350},
  year={2025}
}

@inproceedings{bi2025unveiling,
  title={Unveiling visual perception in language models: An attention head analysis approach},
  author={Bi, Jing and Guo, Junjia and Tang, Yunlong and Wen, Lianggong Bruce and Liu, Zhang and Wang, Bingjie and Xu, Chenliang},
  booktitle={Proceedings of the Computer Vision and Pattern Recognition Conference},
  pages={4135--4144},
  year={2025}
}

@article{zhu2025internvl3,
  title={Internvl3: Exploring advanced training and test-time recipes for open-source multimodal models},
  author={Zhu, Jinguo and Wang, Weiyun and Chen, Zhe and Liu, Zhaoyang and Ye, Shenglong and Gu, Lixin and Tian, Hao and Duan, Yuchen and Su, Weijie and Shao, Jie and others},
  journal={arXiv preprint arXiv:2504.10479},
  year={2025}
}

@article{yang2025qwen3,
  title={Qwen3 technical report},
  author={Yang, An and Li, Anfeng and Yang, Baosong and Zhang, Beichen and Hui, Binyuan and Zheng, Bo and Yu, Bowen and Gao, Chang and Huang, Chengen and Lv, Chenxu and others},
  journal={arXiv preprint arXiv:2505.09388},
  year={2025}
}

@misc{lin2023vila,
      title={VILA: On Pre-training for Visual Language Models},
      author={Ji Lin and Hongxu Yin and Wei Ping and Yao Lu and Pavlo Molchanov and Andrew Tao and Huizi Mao and Jan Kautz and Mohammad Shoeybi and Song Han},
      year={2023},
      eprint={2312.07533},
      archivePrefix={arXiv},
      primaryClass={cs.CV}
}

@inproceedings{pratt2023does,
  title={What does a platypus look like? generating customized prompts for zero-shot image classification},
  author={Pratt, Sarah and Covert, Ian and Liu, Rosanne and Farhadi, Ali},
  booktitle={Proceedings of the IEEE/CVF international conference on computer vision},
  pages={15691--15701},
  year={2023}
}

@inproceedings{alaluf2024myvlm,
  title={Myvlm: Personalizing vlms for user-specific queries},
  author={Alaluf, Yuval and Richardson, Elad and Tulyakov, Sergey and Aberman, Kfir and Cohen-Or, Daniel},
  booktitle={European Conference on Computer Vision},
  pages={73--91},
  year={2024},
  organization={Springer}
}

@article{kuo2022f,
  title={F-vlm: Open-vocabulary object detection upon frozen vision and language models},
  author={Kuo, Weicheng and Cui, Yin and Gu, Xiuye and Piergiovanni, AJ and Angelova, Anelia},
  journal={arXiv preprint arXiv:2209.15639},
  year={2022}
}

@article{chen2025spatial,
  title={Why is spatial reasoning hard for vlms? an attention mechanism perspective on focus areas},
  author={Chen, Shiqi and Zhu, Tongyao and Zhou, Ruochen and Zhang, Jinghan and Gao, Siyang and Niebles, Juan Carlos and Geva, Mor and He, Junxian and Wu, Jiajun and Li, Manling},
  journal={arXiv preprint arXiv:2503.01773},
  year={2025}
}

@article{zhang2025spatial,
  title={Spatial Understanding from Videos: Structured Prompts Meet Simulation Data},
  author={Zhang, Haoyu and Liu, Meng and Li, Zaijing and Wen, Haokun and Guan, Weili and Wang, Yaowei and Nie, Liqiang},
  journal={arXiv preprint arXiv:2506.03642},
  year={2025}
}

@article{li2024topviewrs,
  title={Topviewrs: Vision-language models as top-view spatial reasoners},
  author={Li, Chengzu and Zhang, Caiqi and Zhou, Han and Collier, Nigel and Korhonen, Anna and Vuli{\'c}, Ivan},
  journal={arXiv preprint arXiv:2406.02537},
  year={2024}
}

@inproceedings{li2025llava,
  title={Llava-st: A multimodal large language model for fine-grained spatial-temporal understanding},
  author={Li, Hongyu and Chen, Jinyu and Wei, Ziyu and Huang, Shaofei and Hui, Tianrui and Gao, Jialin and Wei, Xiaoming and Liu, Si},
  booktitle={Proceedings of the Computer Vision and Pattern Recognition Conference},
  pages={8592--8603},
  year={2025}
}

@article{feng2503video,
  title={Video-r1: Reinforcing video reasoning in mllms, 2025},
  author={Feng, Kaituo and Gong, Kaixiong and Li, Bohao and Guo, Zonghao and Wang, Yibing and Peng, Tianshuo and Wu, Junfei and Zhang, Xiaoying and Wang, Benyou and Yue, Xiangyu},
  journal={URL https://arxiv. org/abs/2503.21776}
}

@inproceedings{ma2025spatialllm,
  title={Spatialllm: A compound 3d-informed design towards spatially-intelligent large multimodal models},
  author={Ma, Wufei and Ye, Luoxin and de Melo, Celso M and Yuille, Alan and Chen, Jieneng},
  booktitle={Proceedings of the Computer Vision and Pattern Recognition Conference},
  pages={17249--17260},
  year={2025}
}

@article{he2025spatial,
  title={Spatial-ORMLLM: Improve Spatial Relation Understanding in the Operating Room with Multimodal Large Language Model},
  author={He, Peiqi and Zhang, Zhenhao and Zhang, Yixiang and Zhao, Xiongjun and Peng, Shaoliang},
  journal={arXiv preprint arXiv:2508.08199},
  year={2025}
}

@article{wen2024can,
  title={Can Transformers Capture Spatial Relations between Objects?},
  author={Wen, Chuan and Jayaraman, Dinesh and Gao, Yang},
  journal={arXiv preprint arXiv:2403.00729},
  year={2024}
}

@article{zhang2025mllms,
  title={Why do mllms struggle with spatial understanding? a systematic analysis from data to architecture},
  author={Zhang, Wanyue and Huang, Yibin and Xu, Yangbin and Huang, JingJing and Zhi, Helu and Ren, Shuo and Xu, Wang and Zhang, Jiajun},
  journal={arXiv preprint arXiv:2509.02359},
  year={2025}
}

@article{zheng2025multimodal,
  title={Multimodal Spatial Reasoning in the Large Model Era: A Survey and Benchmarks},
  author={Zheng, Xu and Dongfang, Zihao and Jiang, Lutao and Zheng, Boyuan and Guo, Yulong and Zhang, Zhenquan and Albanese, Giuliano and Yang, Runyi and Ma, Mengjiao and Zhang, Zixin and others},
  journal={arXiv preprint arXiv:2510.25760},
  year={2025}
}

@article{induction,
  title={In-context learning and induction heads},
  author={Olsson, Catherine and Elhage, Nelson and Nanda, Neel and Joseph, Nicholas and DasSarma, Nova and Henighan, Tom and Mann, Ben and Askell, Amanda and Bai, Yuntao and Chen, Anna and others},
  journal={arXiv preprint arXiv:2209.11895},
  year={2022}
}

@article{safety,
  title={On the Role of Attention Heads in Large Language Model Safety},
  author={Zhou, Zhenhong and Yu, Haiyang and Zhang, Xinghua and Xu, Rongwu and Huang, Fei and Wang, Kun and Liu, Yang and Fang, Junfeng and Li, Yongbin},
  journal={arXiv preprint arXiv:2410.13708},
  year={2024}
}

@article{truthful,
  title={Inference-time intervention: Eliciting truthful answers from a language model},
  author={Li, Kenneth and Patel, Oam and Vi{\'e}gas, Fernanda and Pfister, Hanspeter and Wattenberg, Martin},
  journal={Advances in Neural Information Processing Systems},
  volume={36},
  pages={41451--41530},
  year={2023}
}

@article{wu2404retrieval,
  title={Retrieval Head Mechanistically Explains Long-Context Factuality, 2024},
  author={Wu, Wenhao and Wang, Yizhong and Xiao, Guangxuan and Peng, Hao and Fu, Yao},
  journal={URL https://arxiv. org/abs/2404},
  volume={15574}
}

@article{zheng2409attention,
  title={Attention heads of large language models: A survey. arXiv 2024},
  author={Zheng, Z and Wang, Y and Huang, Y and Song, S and Yang, M and Tang, B and Xiong, F and Li, Z},
  journal={arXiv preprint arXiv:2409.03752}
}

@inproceedings{li2020does,
  title={What does BERT with vision look at?},
  author={Li, Liunian Harold and Yatskar, Mark and Yin, Da and Hsieh, Cho-Jui and Chang, Kai-Wei},
  booktitle={Proceedings of the 58th annual meeting of the association for computational linguistics},
  pages={5265--5275},
  year={2020}
}

@inproceedings{cao2020behind,
  title={Behind the scene: Revealing the secrets of pre-trained vision-and-language models},
  author={Cao, Jize and Gan, Zhe and Cheng, Yu and Yu, Licheng and Chen, Yen-Chun and Liu, Jingjing},
  booktitle={European Conference on Computer Vision},
  pages={565--580},
  year={2020},
  organization={Springer}
}

@book{anderson2014rules,
  title={Rules of the mind},
  author={Anderson, John R},
  year={2014},
  publisher={Psychology Press}
}

@article{diamond2013executive,
  title={Executive functions},
  author={Diamond, Adele},
  journal={Annual review of psychology},
  volume={64},
  number={1},
  pages={135--168},
  year={2013},
  publisher={Annual Reviews}
}

@article{cotabish2024spatial,
  title={Spatial Reasoning Excellence: A Synergy of VanTassel-Baska’s Integrated Curriculum Model and Talent Development},
  author={Cotabish, Alicia and Dailey, Debbie and Trumble, Jason and Miller, Rachelle},
  journal={Education Sciences},
  volume={14},
  number={7},
  pages={716},
  year={2024},
  publisher={MDPI}
}

@inproceedings{llm_annotate,
  author       = {Xinru Wang and
                  Hannah Kim and
                  Sajjadur Rahman and
                  Kushan Mitra and
                  Zhengjie Miao},
  editor       = {Florian 'Floyd' Mueller and
                  Penny Kyburz and
                  Julie R. Williamson and
                  Corina Sas and
                  Max L. Wilson and
                  Phoebe O. Toups Dugas and
                  Irina Shklovski},
  title        = {Human-LLM Collaborative Annotation Through Effective Verification
                  of {LLM} Labels},
  booktitle    = {Proceedings of the {CHI} Conference on Human Factors in Computing
                  Systems, {CHI} 2024, Honolulu, HI, USA, May 11-16, 2024},
  pages        = {303:1--303:21},
  publisher    = {{ACM}},
  year         = {2024},
  url          = {https://doi.org/10.1145/3613904.3641960},
  doi          = {10.1145/3613904.3641960},
  bibsource    = {dblp computer science bibliography, https://dblp.org}
}

@article{alain2016understanding,
  title={Understanding intermediate layers using linear classifier probes},
  author={Alain, Guillaume and Bengio, Yoshua},
  journal={arXiv preprint arXiv:1610.01644},
  year={2016}
}

@article{belinkov2022probing,
  title={Probing classifiers: Promises, shortcomings, and advances},
  author={Belinkov, Yonatan},
  journal={Computational Linguistics},
  volume={48},
  number={1},
  pages={207--219},
  year={2022},
  publisher={MIT Press One Broadway, 12th Floor, Cambridge, Massachusetts 02142, USA~…}
}

@article{tenney2019bert,
  title={BERT rediscovers the classical NLP pipeline},
  author={Tenney, Ian and Das, Dipanjan and Pavlick, Ellie},
  journal={arXiv preprint arXiv:1905.05950},
  year={2019}
}

@inproceedings{liu2025feasibility,
  title={Feasibility and Usability Practice on Local Hosting Open Source Large Language Models (LLMs) Including Llama 3.2 Vision 90B in Multi-Functional Agentic Artificial Intelligence (AI) System to Drive Service for Design in the Latest Affordable Small Personal Computer (PC) System},
  author={Liu, Zhen and Wang, Qixu and Lu, Hao and Wang, Yifang},
  booktitle={International Conference on Human-Computer Interaction},
  pages={261--279},
  year={2025},
  organization={Springer}
}

@inproceedings{papineni2002bleu,
  title={Bleu: a method for automatic evaluation of machine translation},
  author={Papineni, Kishore and Roukos, Salim and Ward, Todd and Zhu, Wei-Jing},
  booktitle={Proceedings of the 40th annual meeting of the Association for Computational Linguistics},
  pages={311--318},
  year={2002}
}

@inproceedings{chin2004rouge,
  title={Rouge: A package for automatic evaluation of summaries},
  author={Chin-Yew, Lin},
  booktitle={Proceedings of the Workshop on Text Summarization Branches Out, 2004},
  year={2004}
}

@article{balestri2025gender,
  title={Gender and content bias in Large Language Models: a case study on Google Gemini 2.0 Flash Experimental},
  author={Balestri, Roberto},
  journal={Frontiers in Artificial Intelligence},
  volume={8},
  pages={1558696},
  year={2025},
  publisher={Frontiers Media SA}
}

@article{Liu2022VisualSR,
  title={Visual Spatial Reasoning},
  author={Fangyu Liu and Guy Edward Toh Emerson and Nigel Collier},
  journal={ArXiv},
  year={2022},
  volume={abs/2205.00363}
}

@inproceedings{wang2025spatial457,
  title={Spatial457: A Diagnostic Benchmark for 6D Spatial Reasoning of Large Mutimodal Models},
  author={Wang, Xingrui and Ma, Wufei and Zhang, Tiezheng and de Melo, Celso M and Chen, Jieneng and Yuille, Alan},
  booktitle={Proceedings of the Computer Vision and Pattern Recognition Conference},
  pages={24669--24679},
  year={2025}
}

@article{wang2024picture,
  title={Is a picture worth a thousand words? delving into spatial reasoning for vision language models},
  author={Wang, Jiayu and Ming, Yifei and Shi, Zhenmei and Vineet, Vibhav and Wang, Xin and Li, Sharon and Joshi, Neel},
  journal={Advances in Neural Information Processing Systems},
  volume={37},
  pages={75392--75421},
  year={2024}
}

@inproceedings{ma20253dsrbench,
  title={3dsrbench: A comprehensive 3d spatial reasoning benchmark},
  author={Ma, Wufei and Chen, Haoyu and Zhang, Guofeng and Chou, Yu-Cheng and Chen, Jieneng and de Melo, Celso and Yuille, Alan},
  booktitle={Proceedings of the IEEE/CVF International Conference on Computer Vision},
  pages={6924--6934},
  year={2025}
}

@article{zhang2025look,
  title={Mllms know where to look: Training-free perception of small visual details with multimodal llms},
  author={Zhang, Jiarui and Khayatkhoei, Mahyar and Chhikara, Prateek and Ilievski, Filip},
  journal={arXiv preprint arXiv:2502.17422},
  year={2025}
}

@inproceedings{yin2025spatial,
  title={Spatial mental modeling from limited views},
  author={Yin, Baiqiao and Wang, Qineng and Zhang, Pingyue and Zhang, Jianshu and Wang, Kangrui and Wang, Zihan and Zhang, Jieyu and Chandrasegaran, Keshigeyan and Liu, Han and Krishna, Ranjay and others},
  booktitle={Structural Priors for Vision Workshop at ICCV'25},
  year={2025}
}

@article{cheng2024spatialrgpt,
  title={Spatialrgpt: Grounded spatial reasoning in vision-language models},
  author={Cheng, An-Chieh and Yin, Hongxu and Fu, Yang and Guo, Qiushan and Yang, Ruihan and Kautz, Jan and Wang, Xiaolong and Liu, Sifei},
  journal={Advances in Neural Information Processing Systems},
  volume={37},
  pages={135062--135093},
  year={2024}
}

@article{kravitz2013ventral,
  title={The ventral visual pathway: an expanded neural framework for the processing of object quality},
  author={Kravitz, Dwight J and Saleem, Kadharbatcha S and Baker, Chris I and Ungerleider, Leslie G and Mishkin, Mortimer},
  journal={Trends in cognitive sciences},
  volume={17},
  number={1},
  pages={26--49},
  year={2013},
  publisher={Elsevier}
}

@techreport{openai_o3_o4mini_system_card_2025,
  title        = {OpenAI o3 and o4-mini System Card},
  author       = {{OpenAI}},
  year         = {2025},
  month        = {4},
  day          = {16},
  institution  = {OpenAI},
  url          = {https://cdn.openai.com/pdf/2221c875-02dc-4789-800b-e7758f3722c1/o3-and-o4-mini-system-card.pdf}
}

@article{comanici2025gemini,
  title={Gemini 2.5: Pushing the frontier with advanced reasoning, multimodality, long context, and next generation agentic capabilities},
  author={Comanici, Gheorghe and Bieber, Eric and Schaekermann, Mike and Pasupat, Ice and Sachdeva, Noveen and Dhillon, Inderjit and Blistein, Marcel and Ram, Ori and Zhang, Dan and Rosen, Evan and others},
  journal={arXiv preprint arXiv:2507.06261},
  year={2025}
}
}


\clearpage
\setcounter{page}{1}
\maketitlesupplementary

\subsection{Attention Heat Map for other models}

\begin{figure*}[]
    \centering
    \includegraphics[width=1\textwidth]{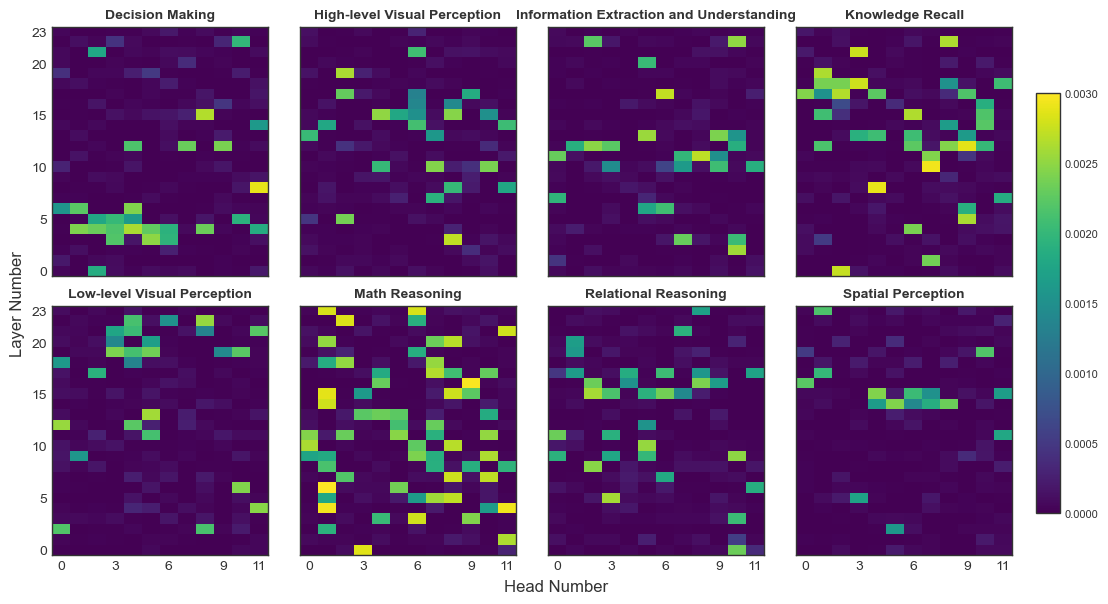}
    \caption{The existence of cognitive heads in Intern3VL-2B. The x-axis represents the head index, while the y-axis indicates
the layer index.}
    \label{app:interven2B}
\end{figure*}

\begin{figure*}
    \centering
    \includegraphics[width=\textwidth]{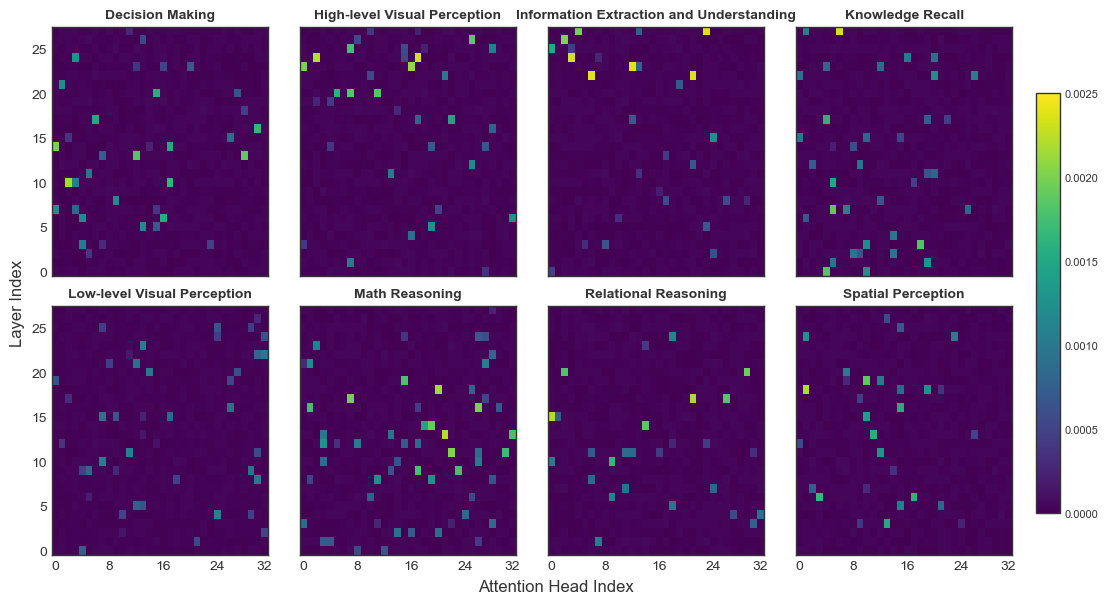}
    \caption{The existence of cognitive heads in LLaMA3.2-11B. The x-axis represents the head index, while the y-axis indicates
the layer index.}
    \label{app:llama11B}
\end{figure*}

\begin{figure*}
    \centering
    \includegraphics[width=1\textwidth]{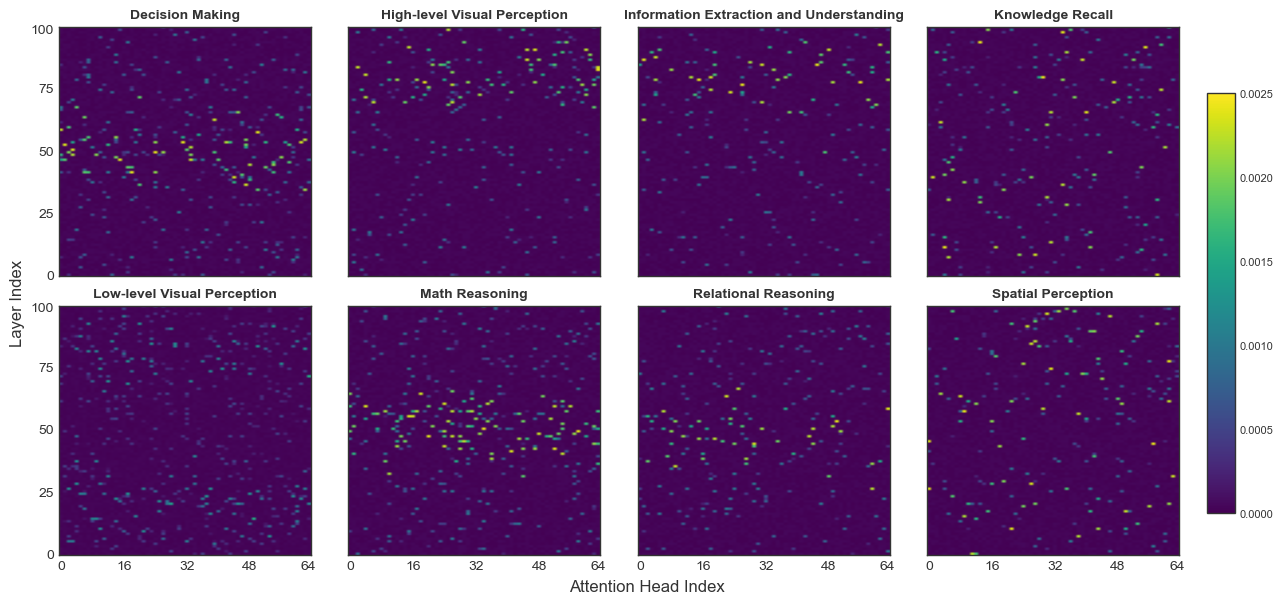}
    \caption{The existence of cognitive heads in LLaMA3.2-90B. The x-axis represents the head index, while the y-axis indicates
the layer index.}
    \label{app:llama90B}
\end{figure*}

\begin{figure*}
    \centering
    \includegraphics[width=1\textwidth]{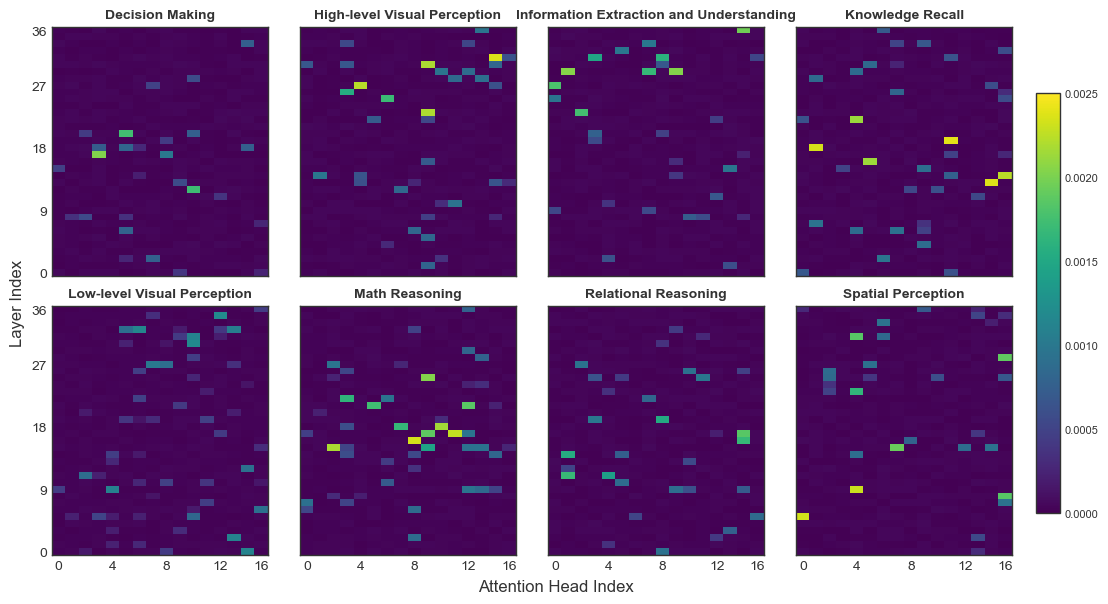}
    \caption{The existence of cognitive heads in Qwen2.5VL-3B. The x-axis represents the head index, while the y-axis indicates
the layer index.}
    \label{app:qwen3B}
\end{figure*}

\begin{figure*}
    \centering
    \includegraphics[width=1\linewidth]{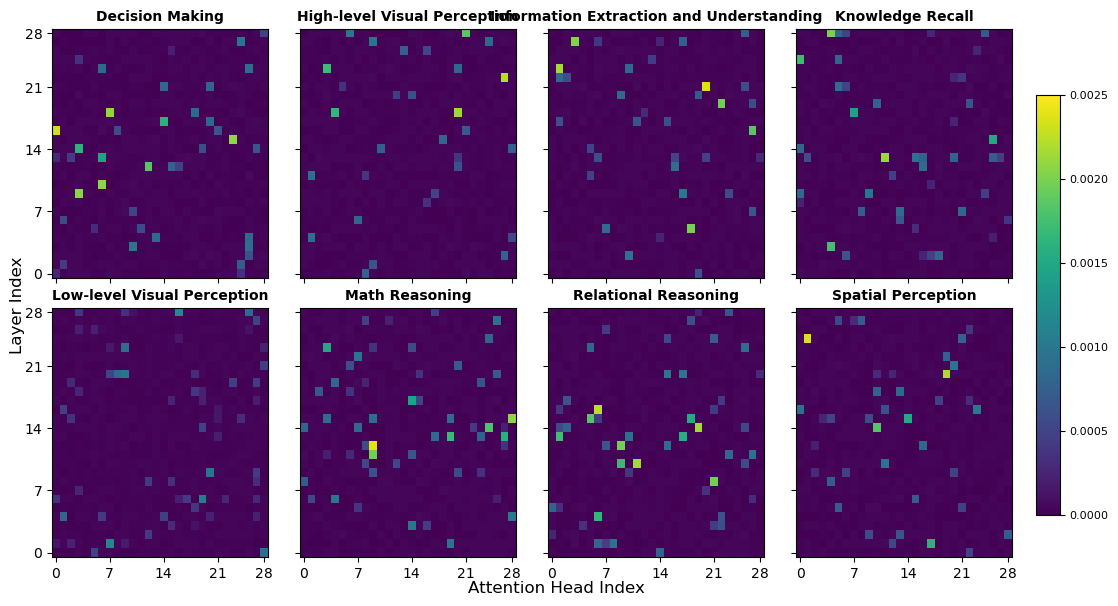}
    \caption{The existence of cognitive heads in Intern3VL-8B. The x-axis represents the head index, while the y-axis indicates
the layer index.}
    \label{app:intern8B}
\end{figure*}

We present the heatmaps \ref{app:interven2B} \ref{app:llama11B} \ref{app:llama90B} \ref{app:qwen3B} \ref{app:intern8B} for the remaining five models in this subsection. The results reveal a
notable universality in the sparsity patterns of attention heads across different architectures.

\subsection{Prompts for Chain-of-Thought Generation}

Below is the specific prompt used to generate the Chain-of-Thought reasoning for the visual spatial reasoning tasks.

\begin{tcolorbox}[
    colback=gray!5!white,       
    colframe=black!75!white,   
    title=\textbf{Generating CoT Prompt Template}, 
    fonttitle=\bfseries,
    sharp corners,
    boxrule=0.8pt
]
    \textbf{Role Definition:} \\
    You are an expert visual spatial reasoning assistant. Given the following question and the correct answer, provide a detailed step-by-step chain of thought reasoning that leads to the correct answer.
    
    \vspace{0.3cm}
    
    \textbf{Input Context:} \\
    \textbf{Question:} \{question\} \\
    \textbf{Correct Answer:} \{correct answer\}
    
    \vspace{0.3cm}
    
    \textbf{Task Instructions:} \\
    Please provide a detailed step-by-step analysis of how to solve this problem. Your response should:
    \begin{enumerate}[leftmargin=2em, noitemsep, topsep=2pt] 
        \item Analyze what you see in the image.
        \item Break down the problem into logical steps.
        \item Conclude with the correct answer.
    \end{enumerate}
    
    \vspace{0.3cm}
    
    \textbf{Output Constraints:} \\
    Format your response as a clear paragraph, each step is one sentence.
\end{tcolorbox}

\subsection{Subquestion Generation Prompt}

To enable more granular control over the reasoning process, we employ a specific prompt designed to decompose complex queries into manageable sub-steps. The prompt explicitly instructs the model to map each step to a specific cognitive skill while adhering to a strict JSON output format. The full prompt content is provided below.

\begin{tcolorbox}[
    colback=gray!5!white,
    colframe=black!75!white,
    title=\textbf{Generating Subquestion Prompt}, 
    fonttitle=\bfseries,
    sharp corners,
    boxrule=0.8pt,
    fontupper=\small    
]
    \textbf{Input Data:} \\
    Here is the question: \\
    \textless question\textgreater \\
    \{question\} \\
    \textless question\textgreater
    
    \par\smallskip 
    
    Here is the chain-of-thought: \\
    \textless chain-of-thought\textgreater \\
    \{cot\} \\
    \textless chain-of-thought\textgreater
    
    \par\smallskip
    
    \textbf{Note:} 
    \begin{itemize}[leftmargin=1.5em, nosep]
        \item Your task is to break the question down into detailed subquestions, ensuring each subquestion can be answered using only one specific cognitive skill.
        \item You need to create a structured and explicit reasoning process that simulates critical thinking while maintaining clarity and precision.
        \item The subquestion needs to be easy to answer and the answer needs to be concise.
        \item The information of chain-of-thought cannot be used directly if it doesn't exist in main query.
        \item Each subquestion should be derived solely from the main query and the preceding subquestion.
        \item You CAN NOT retrieval information from chain-of-thought, but you can retrieval from question.
        \item Each subquestion should be designed to map to exactly one coginitive skill.
        \item Your output should be formatted as a list of JSON objects, where each object represents a subquestion, its answer, and the required cognitive skill.
        \item You should use the most efficient logic to analyze the problem and minimize the number of subquestions.
    \end{itemize}
    
    \par\smallskip
    
    \textbf{Output format:}
    \begin{lstlisting}[basicstyle=\ttfamily\footnotesize, breaklines=true, frame=none, aboveskip=0pt, belowskip=0pt]
[
  {
    "subquestion": "<Subquestion text>",
    "answer": "<Concise answer>",
    "cognitive_skill": "<Assigned cognitive skill>"
  },
  {
    "subquestion": "<Subquestion text>",
    "answer": "<Concise answer>",
    "cognitive_skill": "<Assigned cognitive skill>"
  }
]
    \end{lstlisting}
    
    \textbf{Your answer:}
\end{tcolorbox}

\vspace{5cm}
\subsection{Final Answer Generation Prompt}

Once the reasoning steps are complete, we utilize a strict prompt to consolidate the information and produce the final output. This prompt is designed to enforce a specific JSON format and prevent the model from adding extraneous commentary.

\begin{tcolorbox}[
    colback=gray!5!white,
    colframe=black!75!white,
    title=\textbf{Generating Answer Prompt}, 
    fonttitle=\bfseries,
    sharp corners,
    boxrule=0.8pt,
    fontupper=\small
]
    \textbf{Input Data:} \\
    Here is the question: \\
    \textless question\textgreater \\
    \{question\} \\
    \textless question\textgreater
    
    \par\smallskip
    
    Here is the prior knowledge in chain-of-thought (CoT) format: \\
    \textless prior\_knowledge\textgreater \\
    \{cot\} \\
    \textless prior\_knowledge\textgreater
    
    \par\smallskip
    
    \textbf{Instructions:} 
    \begin{itemize}[leftmargin=1.5em, nosep]
        \item Answer the question carefully.
        \item You can use the information in prior\_knowledge to help you answer the question.
        \item Your response should be clear and concise.
        \item Do not include any explanation, commentary, or code.
        \item Do not output anything after the closing square bracket.
    \end{itemize}
    
    \par\smallskip
    
    \textbf{Output Format:} \\
    Only output your final answer using this format:
    \begin{lstlisting}[basicstyle=\ttfamily\footnotesize, breaklines=true, frame=none, aboveskip=2pt, belowskip=0pt]
[
  {{"answer": "<Your answer here>"}}
]
    \end{lstlisting}
    
    \textbf{Your answer:}
\end{tcolorbox}

\subsection{Token Generation Sample}
\label{subsec:token_sample}

In this section, we provide a raw output sample demonstrating the token-level analysis. Specifically, we extract the top-5 probability tokens during the reasoning phase to monitor the model's decision boundaries.

\begin{tcolorbox}[
    colback=white,
    colframe=black!75!white,
    title=\textbf{Generated Log with Top 5 Tokens},
    fonttitle=\bfseries,
    sharp corners,
    boxrule=0.8pt,
    fontupper=\small
]
    \begin{center}
        \includegraphics[width=0.6\linewidth]{token.png}
    \end{center}
    
    \vspace{0.1cm}
    
    \textbf{Context:} \\
    Consider the real-world 3D location of the objects. Which object is further away from the camera? (A. grey motorcycle, B. green fruits)
    
    \vspace{0.2cm}
    
    \textbf{Model Output Data:}
    
    \begin{lstlisting}[
        basicstyle=\ttfamily\footnotesize, 
        breaklines=true,  
        columns=fullflexible, 
        frame=none,
        escapechar=@ 
    ]
[
  {
    "subquestion": "Which object appears smaller in the image?",
    "answer": "The green fruits appear smaller.",
    "cognitive_skill": "Relational Reasoning"
  },
  {
    "subquestion": "Using perspective rules, which object is farther?",
    "answer": "The green fruits are farther away.",
    "cognitive_skill": "Relational Reasoning",
    "Top_5_Tokens": [
        @\textbf{\color{blue}"bicycle"}@, 
        @\textbf{\color{blue}"sidewalk"}@, 
        @\textbf{\color{blue}"foreground"}@, 
        @\textbf{\color{blue}"nearest"}@, 
        @\textbf{\color{blue}"camera"}@
    ]
  }
]
    \end{lstlisting}

\end{tcolorbox}

\subsection{Spatial Head Augmentation (SHA) Prompt}
\label{subsec:sha_prompt}

To enhance visual grounding, we propose the Spatial Head Augmentation (SHA) method. We utilize \textit{Gemini 2.5-Flash} to generate precise 2D bounding boxes and object segmentation masks. This process ensures that the model focuses on the specific entities referenced in the textual context.

\begin{tcolorbox}[
    colback=gray!5!white,
    colframe=black!75!white,
    title=\textbf{SHA Generation Prompt}, 
    fonttitle=\bfseries,
    sharp corners,
    boxrule=0.8pt,
    fontupper=\small
]
    \textbf{Input Context:} \\
    Given this image with the following context: \\
    \textbf{Question:} \{question\} \\
    \textbf{Answer:} \{answer\}
    
    \vspace{0.3cm}
    
    \textbf{Task Instructions:} \\
    Detect and provide segmentation masks for all prominent objects explicitly mentioned in the question and answer.
    
    \vspace{0.2cm}
    
    \textbf{Output Format:} \\
    Output a JSON list of segmentation masks where each entry contains:
    \begin{itemize}[leftmargin=2em, nosep]
        \item \texttt{"box\_2d"}: the 2D bounding box as [ymin, xmin, ymax, xmax] normalized to 0-1000
        \item \texttt{"mask"}: the segmentation mask as a base64-encoded PNG
        \item \texttt{"label"}: descriptive text label for the object
    \end{itemize}
    
    \vspace{0.2cm}
    
    \textbf{Constraints:} \\
    Focus on the objects mentioned in the question and answer. Only label the explicitly mentioned object like car, bird, person, beach etc., not "the thing", 'it', 'the opposite one'.
\end{tcolorbox}

\subsection{Masked Ratio vs Performance}

\begin{figure*}[]
    \centering
    \includegraphics[width=1\textwidth]{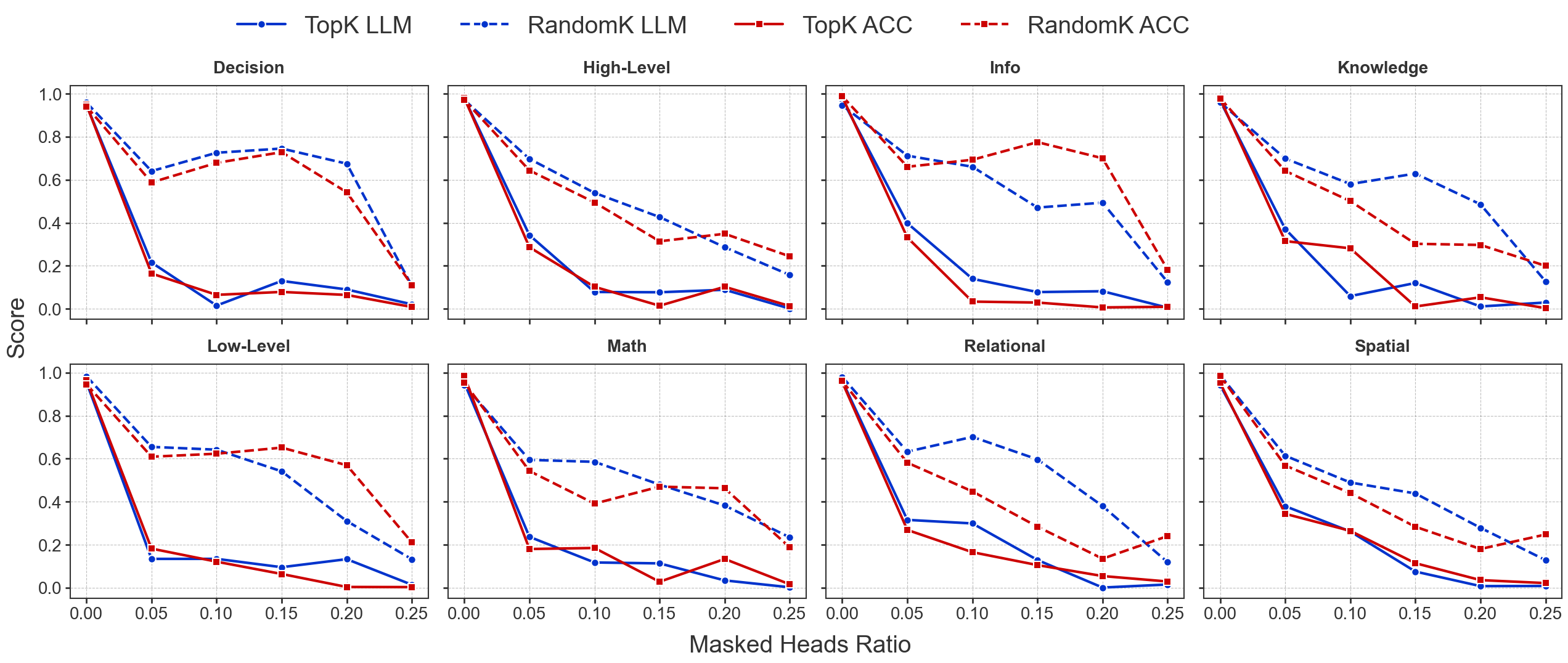}
    \caption{The performance of Qwen2.5-VL-7B after masking out top K cognitive heads vs K random heads on inference, high-level visual reception, low-level visual reception, and decion-masking.}
    \label{fig:Masked Ratio}
\end{figure*}

We gradually mask out the number of cognitive heads and see how the model’s behavior change.
The results can be found in the Figure \ref{fig:Masked Ratio}.

\subsection{Token Position Performance}

We also explore alternative strategies for extracting representations, including using the first token, last token, full tokens. The corresponding results shown in Tale~\ref{tab:token-use-qwen} show that top-k token is the best.

\begin{table*}[t]
\centering
\caption{Performance of different token usage strategies for Qwen2.5-VL-7B. Lower is Better.}
\label{tab:token-use-qwen}
\begin{tabular}{c c c c c}
\toprule
\textbf{Token\_use} & \textbf{Spatial(llm)} & \textbf{Spatial(acc)} & \textbf{Relational(llm)} & \textbf{Relational(acc)} \\
\midrule
first          & 70.42 & 72.54 & 75.28 & 66.46 \\
last           & 84.33 & 83.26 & 78.54 & 70.18 \\
full           & 67.46 & 59.28 & 70.28 & 63.33 \\
top-k          & \textbf{57.69} & \textbf{53.42} & \textbf{63.85} & \textbf{59.57} \\
\bottomrule
\end{tabular}
\end{table*}



\begin{table*}[t]
\centering
\caption{The test accuracy (\%) of probing method on different LLMs.}
\label{tab:probing-accuracy-llms}
\begin{tabular}{lcccccc}
\toprule
\textbf{Dataset} & \textbf{InternVL3-2B} & \textbf{InternVL3-8B} &
\textbf{Qwen2.5VL-7B} & \textbf{Qwen2.5VL-3B} & \textbf{LLaMA3.2-11B} & \textbf{LLaMA3.2-90B} \\
\midrule
CogVSR & 78.74 & 75.88 & 82.15 & 84.22 & 89.64 & 83.44 \\
\bottomrule
\end{tabular}
\end{table*}

\subsection{Case Study in CogVSR}
\label{subsec:case_study}

\begin{figure}[]
    \centering
    \begin{tcolorbox}[
        colback=white,
        colframe=black!75!white,
        title=\textbf{A Concrete Case in CogVSR Dataset},
        fonttitle=\bfseries,
        sharp corners,
        boxrule=0.8pt,
        fontupper=\small
    ]
        \begin{center}
            \includegraphics[width=0.7\linewidth]{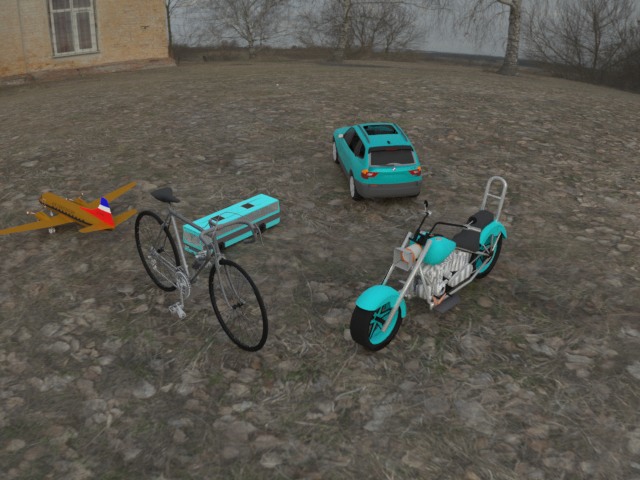}
        \end{center}
        
        \vspace{0.2cm}

        \textbf{Main Question:} \\
        How many planes are left of the blue thing that is in front of the small blue object?
        
        \vspace{0.1cm}
        \textbf{Ground Truth Answer:} 1
        
        \vspace{0.2cm}
        \hrule 
        \vspace{0.2cm}
        
        \textbf{Generated Reasoning Steps:}
        \begin{enumerate}[leftmargin=*, itemsep=0.5em]
            
            \item \textbf{Subquestion:} Which object in the image is the small blue object? \\
            \textit{Answer:} The small blue bus \\
            \textsc{\footnotesize [Skill: High-level Visual Perception]}
            
            \item \textbf{Subquestion:} Which blue thing is located directly in front of that small blue bus? \\
            \textit{Answer:} The big blue car \\
            \textsc{\footnotesize [Skill: Spatial Perception]}

            \item \textbf{Subquestion:} How many planes are located to the left of the big blue car? \\
            \textit{Answer:} 1 \\
            \textsc{\footnotesize [Skill: Math Reasoning]}
            
        \end{enumerate}
    \end{tcolorbox}
    
    \caption{A detailed case in CogVSR dataset.}
    \label{fig:case_study_detail}
\end{figure}

\begin{figure}[]
    \centering
    \begin{tcolorbox}[
        colback=white,
        colframe=black!75!white,
        title=\textbf{A Concrete Case in CogVSR Dataset},
        fonttitle=\bfseries,
        sharp corners,
        boxrule=0.8pt,
        fontupper=\small
    ]
        \begin{center}
            \includegraphics[width=0.7\linewidth]{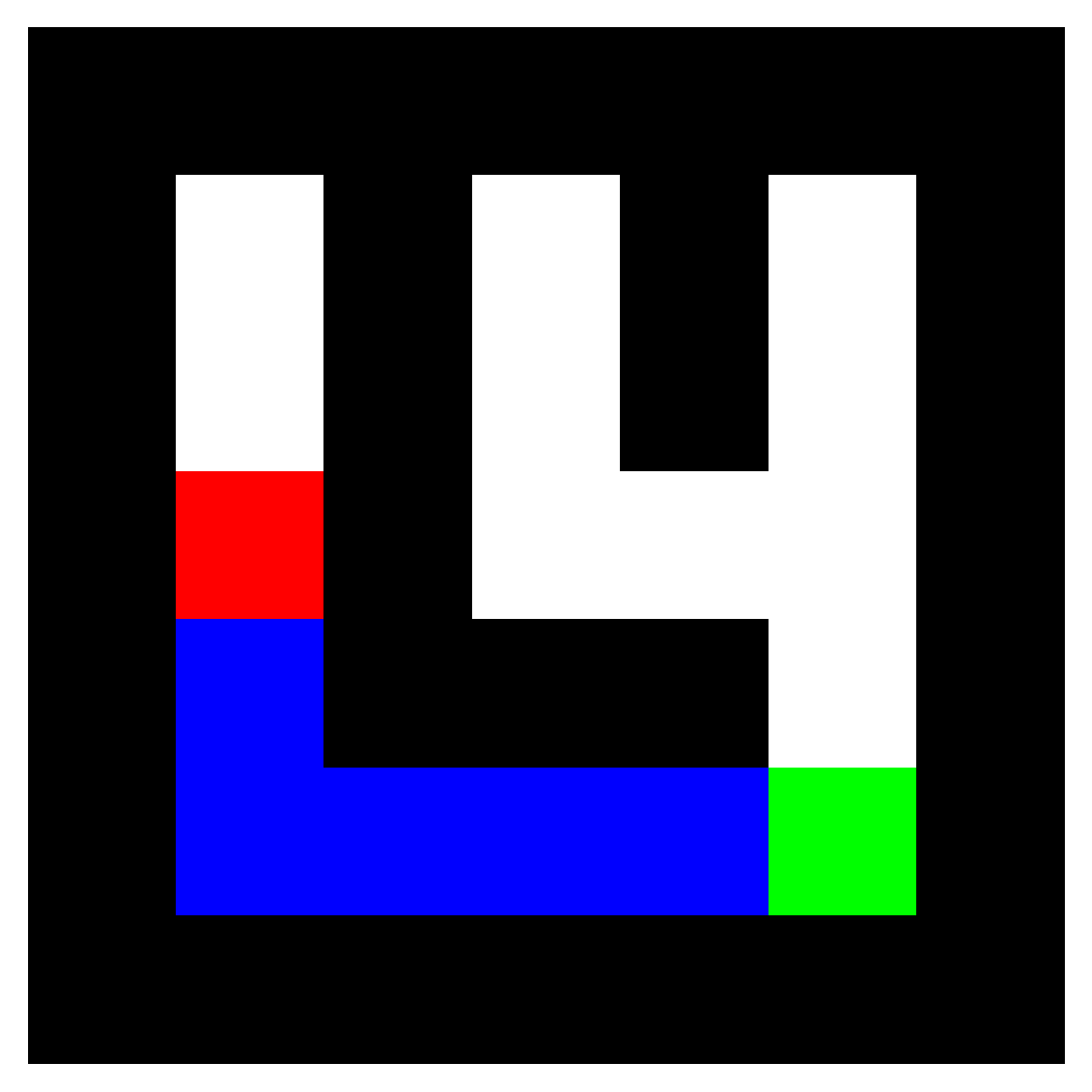}
        \end{center}
        
        \vspace{0.2cm}

        \textbf{Main Question:} \\
        The figure represents a maze where black blocks are walls, white blocks are navigable paths, the green block denotes the starting point (S), and the red block marks the exit (E). The objective is to navigate from S to E by following the path marked in blue, moving only in the four cardinal directions (up, down, left, right). A right turn is defined as a 90° clockwise change in direction, and a left turn as a 90° anticlockwise change. Based on the given blue path, how many right turns occur from S to E? \\[0.3em]
        \textit{Available options:} A. 3 \quad B. 7 \quad C. 4 \quad D. 1.
        
        \vspace{0.1cm}
        \textbf{Ground Truth Answer:} D. 1
        
        \vspace{0.2cm}
        \hrule
        \vspace{0.2cm}
        
        \textbf{Generated Reasoning Steps:}
        \begin{enumerate}[leftmargin=*, itemsep=0.5em]
            
            \item \textbf{Subquestion:} What movement directions does the blue path follow, in sequence, from the starting block (S) to the exit block (E)? \\
            \textit{Answer:} West, then North \\
            \textsc{\footnotesize [Skill: Spatial Perception]}
            
            \item \textbf{Subquestion:} How many times does the path change its movement direction along the blue path from S to E? \\
            \textit{Answer:} 1 \\
            \textsc{\footnotesize [Skill: Relational Reasoning]}
            
            \item \textbf{Subquestion:} Changing direction from West to North corresponds to a 90° clockwise rotation; is this a right turn or a left turn? \\
            \textit{Answer:} Right turn \\
            \textsc{\footnotesize [Skill: Relational Reasoning]}
            
            \item \textbf{Subquestion:} How many right turns does the blue path include from S to E? \\
            \textit{Answer:} 1 \\
            \textsc{\footnotesize [Skill: Math Reasoning]}
            
        \end{enumerate}
    \end{tcolorbox}
    
    \caption{A qualitative example of turn-based maze reasoning. The model first needs to decompose the path into cardinal directions, counts direction changes, classifies the type of turn (right vs.\ left), and finally determines the total number of right turns along the path.}
    \label{fig:case_study_maze_turns}
\end{figure}

\begin{figure}[]
    \centering
    \begin{tcolorbox}[
        colback=white,
        colframe=black!75!white,
        title=\textbf{A Concrete Case in CogVSR Dataset},
        fonttitle=\bfseries,
        sharp corners,
        boxrule=0.8pt,
        fontupper=\small
    ]
        \begin{center}
            \includegraphics[width=0.5\linewidth]{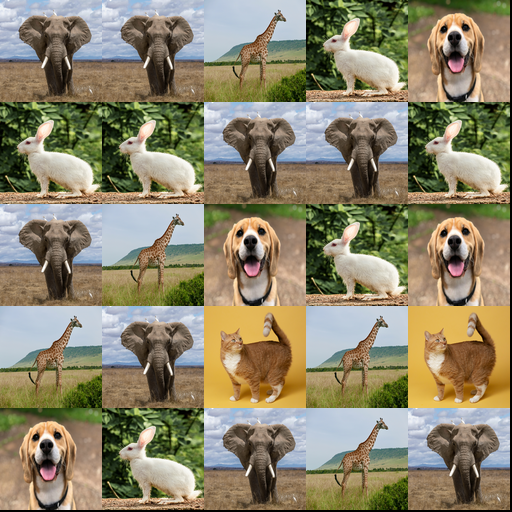}
        \end{center}
        
        \vspace{0.2cm}

        \textbf{Main Question:} \\
        The figure represents a 5$\times$5 grid (5 rows and 5 columns) containing various animals, where each 1$\times$1 square is a block and each block contains an animal from \{\texttt{cat}, \texttt{dog}, \texttt{elephant}, \texttt{giraffe}, \texttt{rabbit}\}. Based on the grid, how many blocks contain a rabbit? \\[0.3em]
        \textit{Available options:} A. 6 \quad B. 5 \quad C. 4 \quad D. 0.
        
        \vspace{0.1cm}
        \textbf{Ground Truth Answer:} A. 6
        
        \vspace{0.2cm}
        \hrule
        \vspace{0.2cm}
        
        \textbf{Generated Reasoning Steps:}
        \begin{enumerate}[leftmargin=*, itemsep=0.5em]
            
            \item \textbf{Subquestion:} How many blocks in row 1 contain a rabbit? \\
            \textit{Answer:} 1 \\
            \textsc{\footnotesize [Skill: High-level Visual Perception]}
            
            \item \textbf{Subquestion:} How many blocks in row 2 contain a rabbit? \\
            \textit{Answer:} 3 \\
            \textsc{\footnotesize [Skill: High-level Visual Perception]}
            
            \item \textbf{Subquestion:} How many blocks in row 3 contain a rabbit? \\
            \textit{Answer:} 1 \\
            \textsc{\footnotesize [Skill: High-level Visual Perception]}
            
            \item \textbf{Subquestion:} How many blocks in row 4 contain a rabbit? \\
            \textit{Answer:} 0 \\
            \textsc{\footnotesize [Skill: High-level Visual Perception]}
            
            \item \textbf{Subquestion:} How many blocks in row 5 contain a rabbit? \\
            \textit{Answer:} 1 \\
            \textsc{\footnotesize [Skill: High-level Visual Perception]}
            
            \item \textbf{Subquestion:} What is the total number of blocks in the entire grid that contain a rabbit? \\
            \textit{Answer:} 6 \\
            \textsc{\footnotesize [Skill: Math Reasoning]}
            
            \item \textbf{Subquestion:} What is the final answer? \\
            \textit{Answer:} A \\
            \textsc{\footnotesize [Skill: Decision]}
            
        \end{enumerate}
    \end{tcolorbox}
    
    \caption{A qualitative example of grid-based counting. The model need to first counts rabbits row by row and then aggregates these subcounts to obtain the total number of rabbit-containing blocks in the grid.}
    \label{fig:case_study_rabbit_grid}
\end{figure}

We present three examples from the test set of the CogVSR dataset in Figure~\ref{fig:case_study_detail}, Figure~\ref{fig:case_study_maze_turns} and Figure~\ref{fig:case_study_rabbit_grid}.

\section{CogVSR functioin details}
\textbf{Spatial Perception}: The capacity to understand the spatial arrangement of objects, including their positions, orientations, distances, and geometric relationships. This function supports tasks involving mental rotation, spatial navigation, and the interpretation of relative location (e.g., left/right, inside/outside).

\textbf{Relational Reasoning}: The capacity to analyze and compare relationships between objects, entities, or concepts (e.g., larger vs. smaller, cause vs. effect). It supports analogical thinking, comparison, and evaluating structural relations rather than isolated features.
    
\textbf{Low-level Visual Perception}: The ability to detect basic visual features such as edges, colors, brightness, motion, and textures, without assigning semantic meaning.

\textbf{High-level Visual Perception}: The ability to recognize and interpret visual content by identifying objects, scenes and semantic meaning.

\textbf{Information Extraction and Understanding}: The ability to identify, retrieve, and semantically interpret relevant information from visual or multimodal inputs.

\textbf{Knowledge Recall}: The ability to retrieve stored facts, concepts, or learned information from long-term memory or pretrained knowledge. It includes recalling definitions, factual associations, and world knowledge relevant to the current task.

\textbf{Math Reasoning}: The ability to process quantitative information, including counting, measuring, estimating magnitude, and performing symbolic or arithmetic operations. It requires integrating numerical rules with logical inference.

\textbf{Decision-Making}: The process of selecting the most appropriate option or answer based on prior reasoning, evaluation of evidence, or predefined objectives.

\section{CogVSR statistics}

The statistics for the CogVSR dataset is shown in Table \ref{app:sta}.

\begin{table}[t]
\centering
\small
\caption{Distribution of cognitive skills across 3{,}759 subquestions.}
\label{tab:cognitive-skill-distribution}
    \resizebox{0.8\linewidth}{!}{
\begin{tabular}{lrr}
\toprule
\textbf{Cognitive Skill} & \textbf{Count} & \textbf{Percentage} \\
\midrule
Spatial Perception                       & 1,026 & 27.29\% \\
High-level Visual Perception             &   941 & 25.03\% \\
Relational Reasoning                     &   576 & 15.32\% \\
Decision Making                          &   498 & 13.25\% \\
Knowledge Recall                         &   272 &  7.24\% \\
Information Extraction and Understanding &   198 &  5.27\% \\
Math Reasoning                           &   138 &  3.67\% \\
Low-level Visual Perception              &   106 &  2.82\% \\
\midrule
\textbf{Total}                           & 3,759 & 100.00\% \\
\bottomrule
\end{tabular}
}
\label{app:sta}
\end{table}



\section{Annotations}\label{app:annotation}

To ensure the quality and reliability of the decomposed subQAF triplets in the CogVision dataset, we design a rigorous multi-stage annotation pipeline, combining expert review and model-based verification. The goal is to verify the logical validity of subquestions, the correctness of their associated cognitive function labels, and the accuracy of the answers.

\paragraph{Stage 1: Validating Subquestion Decomposition}

In the first stage, we evaluate whether the generated subquestions are logically sound and align with natural human reasoning. For each QA pair, three expert annotators (with backgrounds in linguistics or cognitive science) independently assess the validity of each subquestion. A subquestion is marked \texttt{true} if it meaningfully contributes to answering the main question and follows a logical reasoning trajectory. Otherwise, it is marked \texttt{false}.

We apply the following filtering criteria:
\begin{itemize}[noitemsep,topsep=0pt]
    \item \textbf{AI-Human Agreement}: If any annotator considers fewer than 60\% of the subquestions valid, the entire QA decomposition is discarded.
    \item \textbf{Inter-Annotator Agreement}: A subquestion is deemed invalid if at least two annotators mark it as \texttt{false}. If over 40\% of the subquestions in a QA pair are invalid under this rule, the whole QA pair is removed.
\end{itemize}

This filtering ensures that the retained QA decompositions follow coherent, cognitively plausible reasoning chains.

\paragraph{Stage 2: Verifying Cognitive Function Labels}

In the second stage, annotators evaluate the correctness of the function label \(f_i\) assigned to each subQAF triplet \((q_i, a_i, f_i)\). Three annotators independently mark each label as \texttt{true} or \texttt{false}. When discrepancies occur, annotators collaboratively reassign the correct cognitive label to ensure alignment with the underlying mental operation.

This step ensures that the categorization of subquestions accurately reflects established distinctions between information retrieval, semantic understanding, logical reasoning, and other cognitive processes.

\paragraph{Stage 3: Answer Verification via Model and Human Review}

In the final stage, we verify the correctness of each answer \(a_i\) using manual procedure. A human annotator checks and corrects the answer if necessary. Given the generally objective nature of answers, only one annotator is required for this task.



\section{Importance curve}
\label{app-curve}

\begin{figure*}
    \centering
    \includegraphics[width=0.9\textwidth]{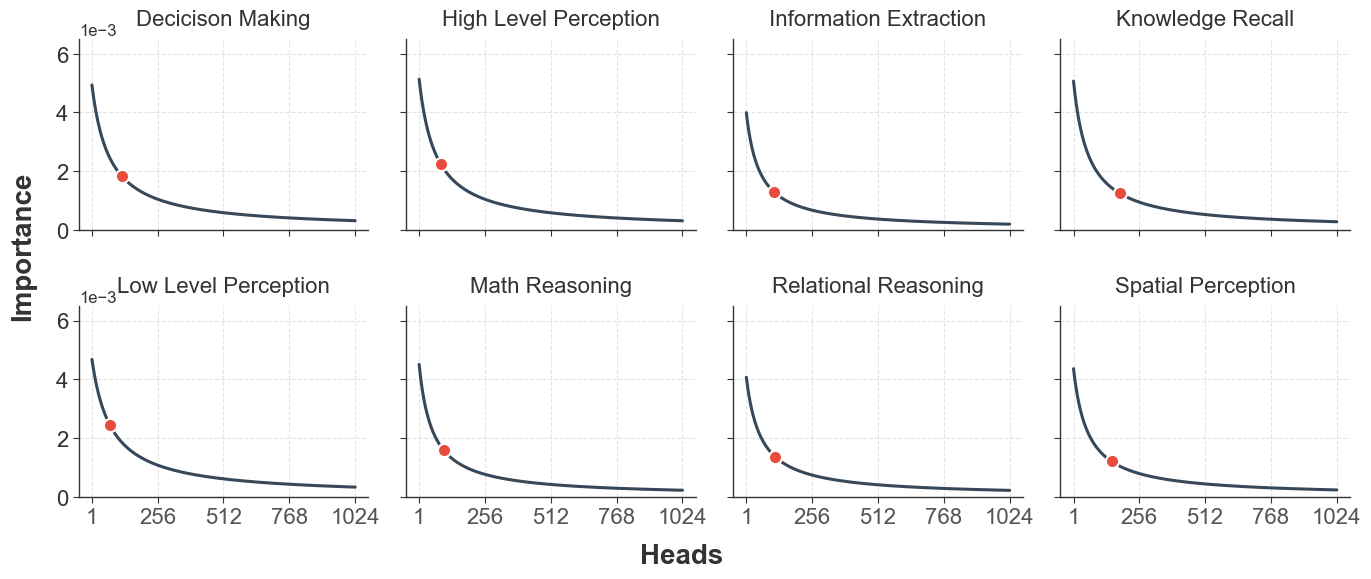}
    \caption{Importance curve for eight functions of Qwen2.5-VL-7B.}
    \label{fig:elbow}
\end{figure*}

We ranked the importance scores and identified the elbow point, as illustrated in Figure~\ref{fig:elbow}.

\section{MLP}\label{mlp}

We train a two-layer multi-class MLP for cognitive function classification. The first layer applies a shared linear projection to each multi-head representation vector, reducing each to a 64-dimensional embedding. These embeddings are then flattened and concatenated into a single vector of size $64\times number of heads$. This vector is fed into a hidden layer with 512 units, followed by a ReLU activation and a dropout with a rate of 0.3. The final output layer maps the 512-dimensional hidden representation to the set of cognitive function labels.

The model is trained using the Adam optimizer with a learning rate of $10^{-4}$ and a cross-entropy loss. Training proceeds for 100 epochs.
The test accuracy of our classification method across all VLM models is summarized in the Table \ref{tab:probing-accuracy-llms}. 







\end{document}